\documentclass[3p]{elsarticle}

\usepackage{amsmath,amssymb,amsfonts}
\usepackage{graphicx}
\usepackage{textcomp}
\usepackage{xcolor}
\usepackage{booktabs}
\usepackage{multirow}
\usepackage{url}
\usepackage{array}
\usepackage{caption}
\usepackage{makecell}
\usepackage{tabularx}
\usepackage{siunitx}
\usepackage{tikz}
\usetikzlibrary{arrows.meta,positioning,calc,shapes.geometric}
\usepackage[colorlinks=true, linkcolor=blue!60!black, citecolor=blue!60!black, urlcolor=blue!60!black]{hyperref}

\newcommand{\dxl}{\Delta x_\ell}
\newcommand{\dyl}{\Delta y_\ell}
\newcommand{\dz}{\Delta z}
\newcommand{\dpsi}{\Delta\psi}

\journal{Robotics and Autonomous Systems}

\makeatletter
\if@twocolumn
  \newenvironment{widefigure}[1][tbp]{\begin{figure*}[#1]}{\end{figure*}}
\else
  \newenvironment{widefigure}[1][tbp]{\begin{figure}[#1]}{\end{figure}}
\fi
\makeatother

\begin{document}

\begin{frontmatter}

\title{WareFly-VLA: A Vision-Language-Action Framework for UAV Navigation and Human Tracking in Smart Warehouses\tnoteref{t1}}
\tnotetext[t1]{Preprint. This manuscript has been submitted to \emph{Robotics and Autonomous Systems} (Elsevier) for possible publication.}

\author[cair]{Thinh D. Le\fnref{eq}}
\ead{thinh.ld@vinuni.edu.vn}
\author[cair]{Son T. Nguyen\fnref{eq}}
\ead{son.nt@vinuni.edu.vn}
\author[cair]{Duong Q. Nguyen}
\ead{duong.nq@vinuni.edu.vn}
\author[cair,cecs]{Dung D. Le}
\ead{dung.ld@vinuni.edu.vn}
\author[cair,cecs,vinrob]{Ngo Anh Vien}
\ead{v.dceo@vinrobotics.net}
\author[cair,cecs]{H. Nguyen-Xuan\corref{cor1}}
\ead{hung.nx2@vinuni.edu.vn}

\affiliation[cair]{organization={Center for AI Research, VinUniversity}, city={Ho Chi Minh City}, country={Vietnam}}
\affiliation[cecs]{organization={College of Engineering and Computer Science, VinUniversity}, city={Hanoi}, country={Vietnam}}
\affiliation[vinrob]{organization={VinRobotics}, city={Hanoi}, country={Vietnam}}

\cortext[cor1]{Corresponding author.}
\fntext[eq]{These authors contributed equally to this work.}

\begin{abstract}
Vision--Language--Action (VLA) models have achieved impressive results in robotic manipulation and ground-mobile navigation, yet language-conditioned control of unmanned aerial vehicles (UAVs) in smart warehouses remains largely unexplored. Progress is hindered by the lack of benchmarks that jointly provide continuous low-level flight actions, fine-grained natural-language target descriptions and realistic industrial environments. This paper introduces WareFly-VLA, a photorealistic UAV VLA framework and dataset for language-guided human search, localization and tracking in intra-logistics and warehouse environments. The dataset contains 507 human-teleoperated flight episodes and 8,504 non-terminal high-resolution RGB transitions collected in NVIDIA Isaac Sim; each transition is paired with a human-written appearance description of the target worker and a synchronized four-degree-of-freedom control command comprising translational and yaw changes. Two aerial embodied tasks are covered, target approach and person following, under realistic challenges including occlusion, long-range search, altitude variation and clutter. A unified benchmark of four representative open-source VLA architectures (SmolVLA, GR00T N1.7, $\pi_0$ and OpenVLA) is established under a leakage-free episode-level protocol at two control rates. The results show that language-conditioned aerial control in warehouses remains far from solved: performance drops substantially under strict generalization settings, continuous action modeling consistently outperforms discrete action tokenization, only the forward channel is reliably learnable from a single frame, and current foundation-model interfaces transfer poorly from ground and humanoid embodiments to aerial platforms. The synchronized video, language, action, pose and difficulty annotations further support world-model research. The dataset, baselines and evaluation protocol are released to support language-grounded aerial autonomy in smart warehouses.
\end{abstract}


\end{frontmatter}

\section{Introduction}

A long-standing objective of embodied artificial intelligence is to enable robots to interpret natural-language instructions, ground them in visual observations and execute appropriate low-level actions in dynamic environments. Recent Vision--Language--Action (VLA) models such as PaLM-E \cite{driess2023palme}, RT-1 \cite{brohan2022rt1}, RT-2 \cite{brohan2023rt2}, Octo \cite{octo2024}, OpenVLA \cite{kim2024openvla}, RDT-1B \cite{liu2024rdt}, $\pi_{0.5}$ \cite{black2025pi05}, SmolVLA \cite{shukor2025smolvla},  TinyVLA \cite{wen2024tinyvla} and Qwen-VLA \cite{qwenvla} have demonstrated that large-scale multimodal training can produce policies that generalize across tasks, environments and robot embodiments. These advances suggest that a unified perception--language--control paradigm may provide a scalable foundation for future autonomous systems.

Despite rapid progress, contemporary VLA research remains heavily concentrated on robotic manipulation and ground-mobile platforms. Large-scale datasets and benchmarks predominantly focus on tabletop manipulation, mobile navigation or humanoid interaction, whereas aerial robots remain significantly underrepresented, even though vision-based control of multi-rotor aerial vehicles is a mature and active research area \cite{nascimento2026review}. This gap is particularly notable because unmanned aerial vehicles (UAVs) present a distinct embodiment with substantially different sensing, motion and control characteristics. Being different from manipulators that operate within constrained workspaces or ground robots that navigate primarily in two dimensions, UAVs perform continuous motion in three-dimensional space while simultaneously regulating position, altitude and heading. Visual observations are highly dynamic due to rapid viewpoint changes, target scale variation and frequent occlusions. Consequently, transferring VLA paradigms developed for ground or manipulation settings to aerial systems is far from straightforward.

Among the many tasks relevant to aerial autonomy, language-guided human search and following represents a particularly compelling benchmark for embodied intelligence. Consider the instruction: \emph{``Track the worker wearing a yellow hard hat and a purple safety vest.''} Successfully executing this command requires the integration of multiple capabilities. The UAV must first interpret fine-grained appearance descriptions, identify the correct individual among visually similar distractors, actively search when the target is not immediately visible, maintain target association over time and continuously generate flight-control commands that ensure safe and reliable tracking. In contrast to conventional UAV tracking benchmarks, where the target is initialized through a bounding box or visual template, language-guided following requires semantic target specification and persistent multimodal grounding throughout the entire trajectory. This setting naturally combines visual grounding, embodied language understanding, active perception and continuous control within a single task.

Existing datasets do not adequately support this problem formulation. Classical approaches such as Chen et al. \cite{chen2018objTracking}, aerial tracking benchmarks such as UAV123 \cite{mueller2016uav}, and subsequent UAV tracking datasets\cite{zhao2022antiUAV} provide extensive flight recordings but rely exclusively on visual target initialization without language supervision. Vision-language navigation datasets such as AerialVLN \cite{liu2023aerialvln}, AirNav \cite{deshpande2026airnav}, EmbodiedCity \cite{zhang2024embodiedcity}, UAV-VLN \cite{wang2024uav} and GeoText-1652 \cite{mei2024geotext} introduced natural-language instructions but formulated navigation primarily as waypoint prediction or route planning rather than continuous low-level control. In a similar vein, large-scale drone footage datasets exist and are used to extract ground vehicle trajectories for driving research\cite{wang2025dadd}; however, the lack of ergo-centric controls makes them unsuitable for aerial embodied AI. Recent UAV-oriented VLA datasets and benchmarks remain limited in scope. CognitiveDrone \cite{cogdrone2025} focused on simulation-based trajectory generation. UAV-VLA \cite{sautenkov2025uavvla} targets mission-level reasoning from aerial imagery and RaceVLA \cite{serpiva2025racevla} addressed drone-racing policies. Although these resources have advanced aerial embodied learning, none simultaneously provide appearance-grounded language descriptions, continuous flight-control supervision, human-demonstrated trajectories and warehouse-scale person-search-and-follow tasks.

To address this limitation, this paper introduces \textbf{WareFly-VLA}, a Vision--Language--Action dataset for language-guided human search and following in industrial warehouse environments. The dataset is collected from \emph{human-teleoperated UAV demonstrations} performed in a photorealistic warehouse environment constructed in NVIDIA Isaac Sim. Each trajectory contains synchronized high-resolution RGB observations, fine-grained appearance-based language descriptions and continuous 4-DoF flight actions represented as body-frame motion commands. In contrast to scripted navigation policies or automatically generated expert trajectories, the dataset captures realistic human decision-making behaviors during target search, target acquisition and continuous following. The resulting trajectories naturally include recovery maneuvers, viewpoint adaptation, target re-identification and other behaviors that commonly arise during practical aerial operation.

WareFly-VLA is designed around two complementary embodied tasks: \emph{approach-and-locate} and \emph{person-following}. The dataset further incorporates diverse warehouse layouts, worker appearances, target distances, altitude changes, partial occlusions and cluttered industrial environments. Structured difficulty annotations enable systematic evaluation across multiple challenge dimensions and support future research on generalization, robustness and language grounding under realistic operating conditions.

Beyond dataset construction, a unified benchmark is established to evaluate representative open-source VLA architectures using a common leakage-free episode-level split. The benchmark spans multiple model families and action-representation paradigms, providing a standardized framework for studying aerial embodied intelligence. Experimental results indicate that language-conditioned UAV control remains a challenging open problem, highlighting substantial opportunities for future advances in multimodal representation learning, continuous action prediction and cross-embodiment transfer.

The main contributions of this work are as follows:

\begin{enumerate}

\item \textbf{WareFly-VLA}, a human-teleoperated UAV Vision--Language--Action dataset for language-guided human search and following in warehouse environments, containing synchronized visual observations, appearance-grounded language descriptions and continuous 4-DoF flight actions.

\item \textbf{A human-demonstration data collection pipeline} built in NVIDIA Isaac Sim that captures realistic aerial search, target localization, target acquisition and continuously following behaviors from teleoperated quadrotor trajectories.

\item \textbf{Aerial embodied benchmark tasks} covering approach-and-locate and person-following scenarios under realistic challenges, including occlusion, altitude variation, long-range search, viewpoint changes and cluttered industrial environments.

\item \textbf{A unified aerial VLA benchmark} for evaluating representative open-source VLA architectures under a standardized leakage-free evaluation protocol.

\item \textbf{An open-source research platform} including dataset artifacts, benchmark configurations, training pipelines and evaluation tools to facilitate future research on aerial embodied intelligence, language-grounded UAV autonomy and Vision--Language--Action learning.

\item \textbf{Open world-model research directions} enabled by WareFly-VLA, including predictive representation learning, action-conditioned latent dynamics, language-conditioned planning and sim-to-real transfer for aerial embodied intelligence.
\end{enumerate}

\section{Related Work}
\label{sec:related}

\subsection{Vision--Language--Action models}

VLA models have emerged as a unified paradigm for embodied intelligence by jointly learning perception, language understanding and control. Early efforts such as PaLM-E \cite{driess2023palme} demonstrated the feasibility of integrating visual observations into large language models for robotic decision making. Subsequent systems including RT-1 \cite{brohan2022rt1}, RT-2 \cite{brohan2023rt2}, Open X-Embodiment \cite{padalkar2023rtx}, Octo \cite{octo2024}, OpenVLA \cite{kim2024openvla}, RDT-1B \cite{liu2024rdt}, $\pi_{0.5}$ \cite{black2025pi05}, SmolVLA \cite{shukor2025smolvla}, TinyVLA \cite{wen2024tinyvla} and Qwen-VLA \cite{qwenvla} have demonstrated increasingly strong generalization across tasks, environments and robotic embodiments.

Recent research has further explored large-scale pretraining and cross-embodiment transfer. Open X-Embodiment \cite{padalkar2023rtx} aggregated demonstrations from multiple robot platforms to investigate shared policy learning, while OpenVLA \cite{kim2024openvla}, RDT-1B \cite{liu2024rdt} and $\pi_0$ \cite{black2024pi0} explored foundation-model approaches capable of adapting to diverse downstream tasks. Inference-time policy steering has further been proposed to deploy pre-trained VLAs on new tasks without fine-tuning \cite{li2026vlapilot}. Despite these advances, the overwhelming majority of VLA benchmarks remain centered on robotic manipulation, mobile navigation and humanoid control. The applicability of VLA models to aerial robots remains comparatively underexplored, largely due to the scarcity of UAV-specific datasets that combine language instructions with continuous flight-control supervision.

\subsection{Language-grounded aerial navigation}

Language-guided navigation has become an important benchmark for embodied intelligence, motivating a growing body of research on aerial vision-language systems. AerialVLN \cite{liu2023aerialvln} introduced a large-scale simulation benchmark for instruction-following UAV navigation. Subsequent efforts expanded this direction through more realistic environments and richer navigation objectives, including UAV-VLN \cite{wang2024uav}, EmbodiedCity \cite{zhang2024embodiedcity}, GeoText-1652 \cite{mei2024geotext}, AirNav \cite{deshpande2026airnav} and semantic-topometric aerial navigation frameworks \cite{gao2024aerial}. Additional studies have investigated aerial dialogue navigation and interactive instruction following \cite{fan2022aerial}, and vision-language policies for onboard drone navigation have been trained in radiance-field simulators \cite{chen2026gradnav}. In parallel, learning-based vision-driven following behaviors for UAVs, such as safe river following via reinforcement learning with a semantic dynamics model \cite{wang2026river}, show that end-to-end visual control of aerial platforms is feasible, but they do not involve language-specified targets.

These datasets have significantly advanced aerial language grounding and instruction-conditioned navigation. However, most methods formulate navigation as waypoint prediction, route planning or localization rather than direct low-level control. As a result, they provide limited supervision for end-to-end VLA learning, where language instructions must be translated directly into continuous flight actions.

\subsection{UAV VLA datasets and benchmarks}

Recent studies have begun extending the Vision--Language--Action paradigm to aerial robotics. UAV-VLA \cite{sautenkov2025uavvla} investigates mission-level planning and reasoning from aerial imagery, while RaceVLA \cite{serpiva2025racevla} explores language-conditioned drone-racing policies derived from OpenVLA. CognitiveDrone \cite{cogdrone2025} combines VLA policies with reasoning-oriented VLMs for simulation-based cognitive UAV navigation. Broader surveys have also highlighted the growing role of foundation models and LLMs in UAV autonomy \cite{tian2025uavllm}.

Although these efforts demonstrate the promise of foundation-model-based aerial autonomy, important limitations remain. Existing benchmarks are typically restricted to racing, navigation, mission planning or simulation-specific tasks. Many datasets rely on automatically generated trajectories or task-specific objectives that do not require semantic target grounding. Furthermore, appearance-based human search and following has received little attention within the UAV VLA literature.

Consequently, there remains no publicly available benchmark that simultaneously provides appearance-grounded natural-language target descriptions, continuous 4-DoF flight-control actions, human-teleoperated UAV demonstrations and language-guided human search-and-follow tasks in realistic warehouse environments. WareFly-VLA is designed to fill this gap and to provide a dedicated benchmark for aerial Vision--Language--Action learning.

\section{The WareFly-VLA dataset}
\label{sec:dataset}

\subsection{Dataset design principles}

WareFly-VLA is designed around a central objective: enabling research on language-guided human search and following for aerial robots. In contrast to conventional UAV tracking benchmarks that require explicit target initialization via bounding boxes or visual templates, WareFly-VLA specifies the target exclusively through natural-language descriptions of observable appearance attributes. Consequently, successful task completion requires the integration of language grounding, visual identification, active search, target re-identification and continuous flight control.

Three design principles guided the dataset construction process. Firstly, all trajectories are generated from \emph{human-teleoperated demonstrations} rather than scripted controllers, allowing the dataset to capture realistic search and following behaviors. Secondly, language annotations emphasize appearance-based target grounding instead of spatial instructions, reflecting realistic warehouse deployment scenarios. Thirdly, the dataset focuses on continuous low-level flight actions, making it suitable for Vision--Language--Action learning rather than high-level navigation alone.

\subsection{Simulation environment}

Data collection is implemented in Pegasus Simulator\cite{pegasus_sim}, a framework for simulating drone dynamics built as an extension to NVIDIA Isaac Sim\cite{NVIDIA_Isaac_Sim}, a photorealistic and physically based robotics simulation platform. The environment consists of a large industrial warehouse containing shelving racks, conveyor systems, forklift lanes, storage areas and mixed lighting conditions. The scene is intentionally designed to expose the UAV to realistic visual challenges, including clutter, repetitive textures, reflective surfaces, viewpoint changes and partial occlusions. A simulated quadrotor equipped with a forward-facing RGB camera captures observations at $1920\times1440$ pixels. Ground-truth vehicle pose is obtained directly from the simulator state, providing accurate position and orientation measurements throughout each trajectory.

\subsection{Human-Teleoperated Data Collection}

Each episode begins by assigning a worker a unique combination of clothing and personal protective equipment (PPE), including safety vests, hard hats, shirts, overalls, pants, gloves and other accessories. Appearance configurations are intentionally diversified to maximize language grounding diversity and minimize attribute overlap between episodes.

A human operator teleoperates the UAV while attempting to locate, approach and follow the designated target worker. The target follows a predefined route unknown to the operator, requiring continuous visual search and target maintenance throughout the trajectory. This setup naturally produces realistic aerial behaviors, including target acquisition, viewpoint adjustment, temporary target loss, recovery maneuvers and continuous following.

All RGB observations, control inputs and simulator states are recorded synchronously. The resulting trajectories form a collection of human-demonstrated aerial embodied behaviors suitable for Vision--Language--Action learning.

To reduce redundancy while preserving meaningful control dynamics, the
recorded streams are temporally subsampled to $1$ FPS, consistent with the effective decision frequencies commonly adopted in recent VLA
systems \cite{black2024pi0,kim2024openvla}.

\subsection{Teleoperation policy}
\label{sec:teleop_policy}

\begin{figure}[tbp]
\centering
\includegraphics[width=\linewidth,height=0.72\textheight,keepaspectratio]{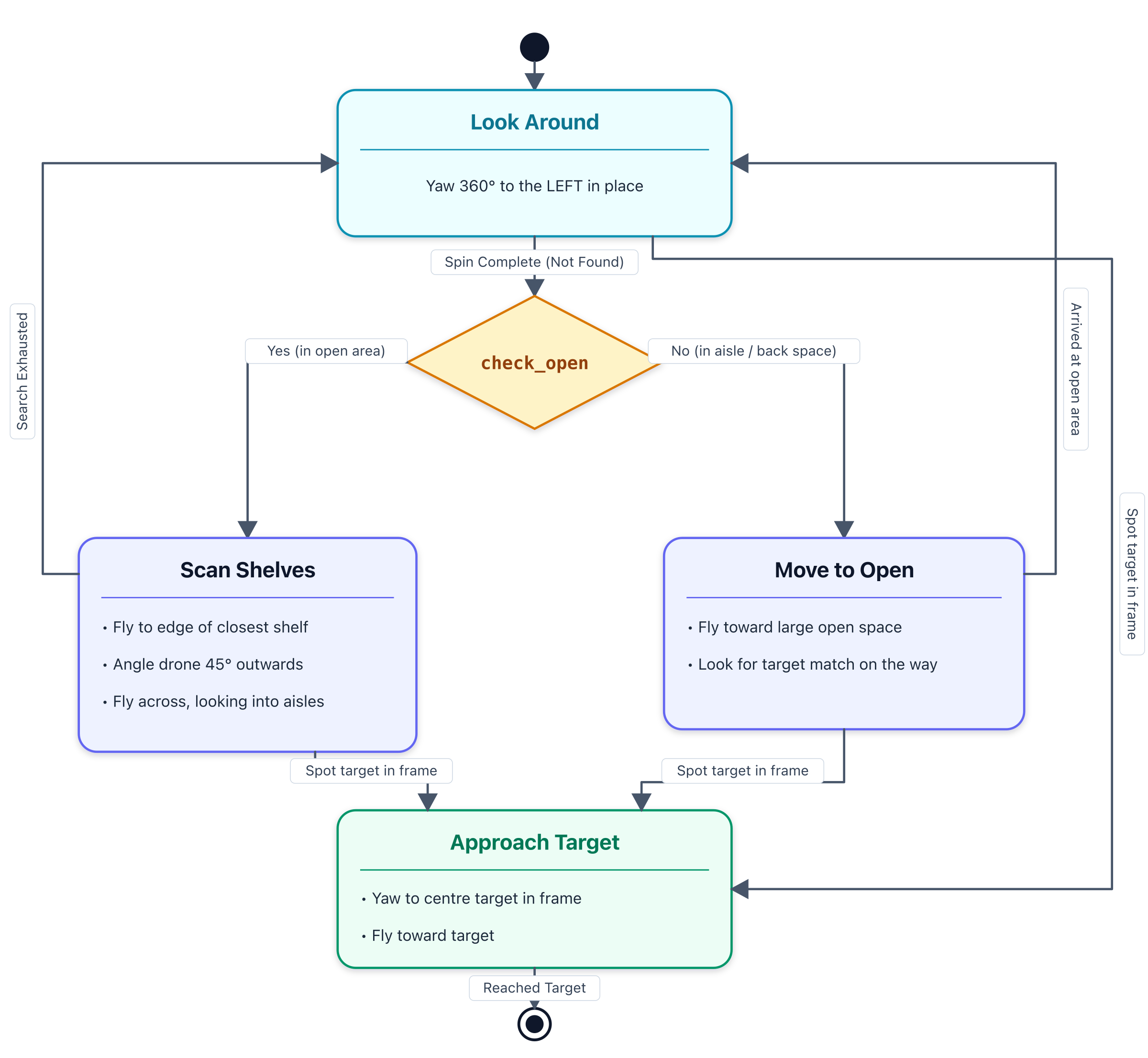}
\caption{State-transition diagram of the human teleoperation policy used during data collection. \emph{Search states} (\emph{Look Around}, \emph{Scan Shelves} and \emph{Move to Open}) are shown in cyan. The \texttt{check\_open} decision node is shown in amber, and the goal-directed \emph{Approach Target} state is shown in green. From any search state, detecting the target in the camera frame triggers an immediate transition to \emph{Approach Target}. If \emph{Scan Shelves} ends with \emph{Search Exhausted}, or if \emph{Move to Open} ends with \emph{Arrived at open area}, the policy returns to \emph{Look Around}. The episode terminates when the drone reaches the target.}
\label{fig:teleop_policy}
\end{figure}

To produce trajectories that are diverse in scene yet consistent in behavior, the human pilot follows a deliberately deterministic \emph{search--decide--approach} policy. Fig. \ref{fig:teleop_policy} summarises the policy as a finite-state machine with three search states, one decision, one goal-directed action state and an absorbing terminal state.

\emph{Look Around.} Every episode begins with the pilot performing a yaw $360^{\circ}$ sweep in place. The sweep is a rotation-only maneuver around the drone's initial position. It enables target acquisition within the camera's effective field of view at the current altitude without committing the drone to a specific motion direction.

\emph{\texttt{check\_open} decision.} If the sweep completes without acquiring the target, the pilot consults a binary state-of-environment check. When the drone is currently in an \emph{open area} (sufficient lateral clearance to fly between shelving rows), control proceeds to \emph{Scan Shelves}. When the drone is in an \emph{aisle or back-space} (constrained by walls or rack ends), control proceeds to \emph{Move to Open}.

\emph{Scan Shelves.} In this state the pilot flies to the edge of the closest shelving row, angles the drone outward by approximately $45^{\circ}$ and traverses parallel to the row while looking down each aisle in turn. This pattern systematically sweeps the only volume that the in-place yaw cannot reach the deep interior of shelving aisles and is the dominant source of large lateral $\dyl$ commands in the corpus.

\emph{Move to Open.} When the drone is hemmed in by infrastructure, the pilot first repositions to a viewpoint with broader lateral clearance, flying toward the nearest open area and continuing to look for an appearance-matching target along the way. Once the drone arrives, the state machine loops back to \emph{Look Around} so that a full sweep can be executed from a more advantageous viewpoint.

\emph{Approach Target.} As soon as the target appears in the camera frame from \emph{any} of the three search states, the pilot transitions to the goal-directed \emph{Approach Target} state, yaws to center the target in the frame, and flies toward it. The episode terminates on \emph{Reached Target}.

\emph{Search-exhaustion loops.} A \emph{Scan Shelves} pass that completes without acquisition raises a \emph{Search Exhausted} transition back to \emph{Look Around} and a \emph{Move to Open} pass that arrives at the new viewpoint without acquisition raises an analogous \emph{Arrived at open area} transition. The policy therefore guarantees that the drone always returns to a yaw-sweep state between unsuccessful search legs, rather than chaining multiple action states blindly.

\emph{Why a deterministic policy.} Fixing the high-level behavior while randomizing warehouse layout, target route and target appearance configuration produces a corpus that is \emph{behaviourally consistent} every episode contains the same acquisition--closure--tracking phases while remaining \emph{trajectory-wise diverse}, with varied path shapes, distances, durations and occlusion patterns (Fig. \ref{fig:trajectories}). This combination is what allows the 507-episode corpus to discriminate sharply between modern VLA architectures despite its modest scale (Sections \ref{sec:experiments}--\ref{sec:results}) and it directly motivates the difficulty-tag axes used for stratified evaluation (Section \ref{sec:analysis}).

\subsection{Action representation}

Continuous flight actions are derived from consecutive simulator poses. Given position $\mathbf{p}_t$ and heading $\psi_t$ at time $t$, the action label is represented as a four-dimensional body-frame motion command

\begin{equation}
\mathbf{a}_t =
(\Delta x,\Delta y,\Delta z,\Delta \psi).
\end{equation}

The translational part is expressed in the UAV body frame:

\begin{equation}
\begin{bmatrix}
\Delta x \\
\Delta y
\end{bmatrix}
=
R(-\psi_t)
\left(
\mathbf{p}_{t+1}^{xy}
-
\mathbf{p}_{t}^{xy}
\right)
\end{equation}
while altitude and heading updates are computed as
\begin{align}
\Delta z &= z_{t+1}-z_t, \\
\Delta \psi &= \mathrm{wrap}_{[-\pi,\pi]}
(\psi_{t+1}-\psi_t).
\end{align}

This representation is invariant to global orientation and directly
compatible with UAV controllers operating in local body coordinates.

\subsection{Language annotation}

Following trajectory collection, each episode is independently annotated by human annotators from the perspective of a remote UAV operator. Instructions describe the target exclusively through observable appearance attributes, including clothing, colors, accessories, and personal protective equipment (PPE), while avoiding identity information or scene-specific positional references. Multiple paraphrases are provided to increase linguistic diversity and improve robustness to instruction variations. All annotations undergo secondary review to ensure factual consistency and language quality, resulting in an appearance-centric language corpus specifically designed for language-guided human search and following.

\section{Dataset analysis}
\label{sec:analysis}

\subsection{Quantitative statistics}

Table \ref{tab:stats} summarizes the key statistics of WareFly-VLA. The dataset contains $507$ episodes and $8{,}504$ non-terminal transitions with valid control actions. All frames are captured at a fixed resolution of $1920\times1440$ pixels. This resolution provides high visual fidelity for fine-grained appearance grounding. The trajectories exhibit substantial drone motion across both short-range maneuvers and long-distance approach episodes. They also show a large variation in pursuit distance and motion pace, reflecting diverse target routes and human teleoperation strategies.

\begin{table}[t]
\renewcommand{\arraystretch}{1.2}
\centering
\caption{WareFly-VLA dataset statistics.}
\label{tab:stats}
\begin{tabular}{lc}
\toprule
\textbf{Property} & \textbf{Value} \\
\midrule
Total episodes              & 507 \\
Non-terminal transitions    & 8{,}504 \\
\quad train / val           & 7{,}276 / 1{,}228 \\
Train / val episodes        & 431 / 76 \\
Frame rate                  & 1 FPS \\
Image resolution            & $1920\times1440$ px \\
Action dimensionality       & 4-DoF $(\dxl,\dyl,\dz,\dpsi)$ \\
Task-verb families          & 15 \\
Instruction vocabulary size & 448 tokens \\
Mean instruction length     & $20.0 \pm 5.1$ words \\
Challenge-tagged episodes   & yes (\emph{find/close/far/height\_change/obstructed}) \\
\bottomrule
\end{tabular}
\end{table}

Fig. \ref{fig:overview} visualizes the episode-length distribution, the per-episode frame-count profile and the task-verb breakdown. Fig. \ref{fig:trajectories} shows representative drone trajectories in the world-frame $xy$-plane together with a top-down composite over the warehouse scene, illustrating the variety of flight paths across episodes and confirming that no two episodes share identical dynamics.

\begin{widefigure}[tbp]
  \centering
  \includegraphics[width=\linewidth,height=0.72\textheight,keepaspectratio]{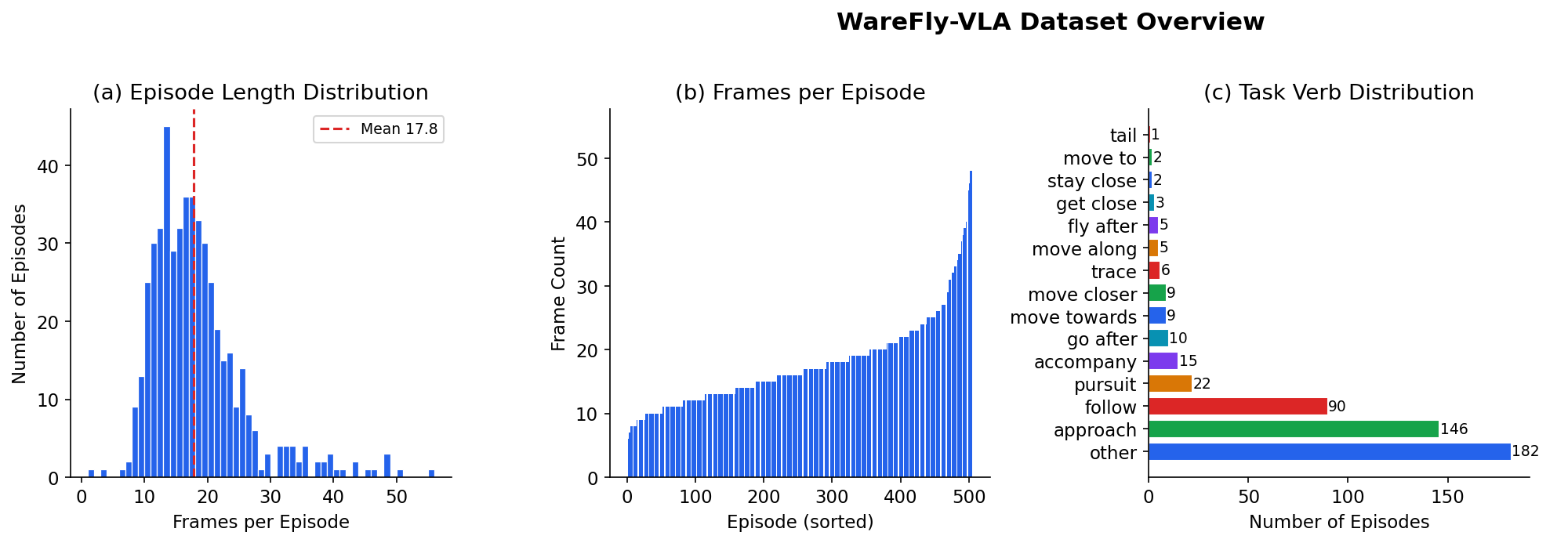}
  \caption{Dataset overview.
    \textbf{(a)}~Distribution of episode lengths;
    the dashed red line marks the mean.
    \textbf{(b)}~Per-episode frame count sorted in ascending order.
    \textbf{(c)}~Task-verb distribution across $15$ phrasings;
    \textit{approach} and \textit{follow} dominate, with thirteen additional verb families ensuring paraphrase diversity.}
  \label{fig:overview}
\end{widefigure}

\begin{widefigure}[tbp]
  \centering
  \includegraphics[width=\linewidth,height=0.72\textheight,keepaspectratio]{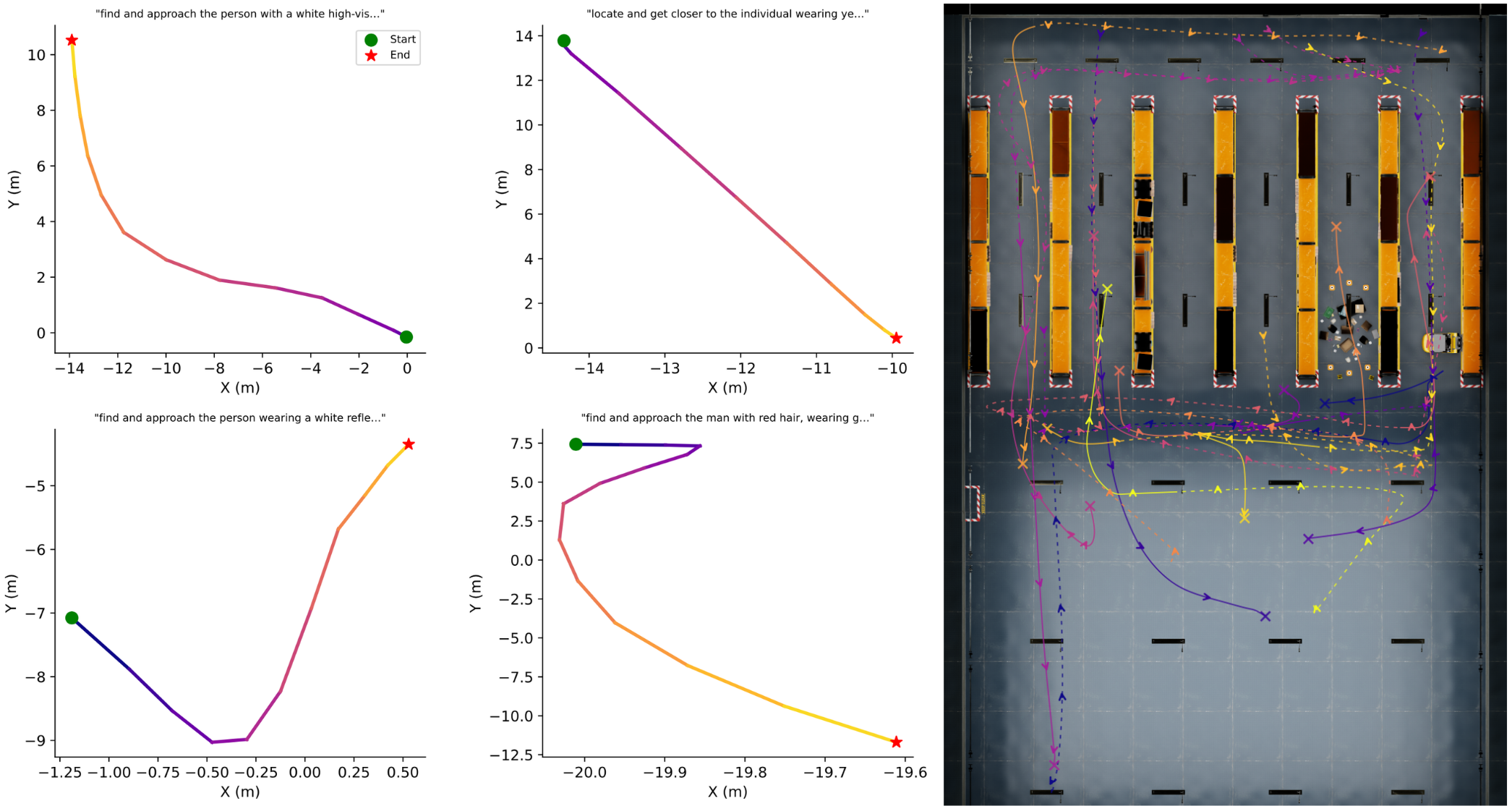}
  \caption{Drone flight trajectories in world coordinates. \textbf{Left:} four representative single-episode trajectories in the world-frame $xy$-plane, color-coded from early (blue/purple) to late (yellow) via the plasma colormap, with the language instruction shown above each panel; green circles indicate episode start positions and red stars indicate end positions. \textbf{Right:} top-down composite of many episode trajectories overlaid on a bird's-eye view of the photorealistic Isaac~Sim warehouse. \emph{Solid line segments} denote frames in which the drone has the target in view, while \emph{dashed line segments} denote frames in which the target is out of view (search / re-acquisition). The composite illustrates the spatial diversity of pursuit paths across shelving rows, aisles and open areas and confirms that the dataset captures a wide range of pursuit behaviors rather than a single stereotyped maneuver.}
  \label{fig:trajectories}
\end{widefigure}

\subsection{Action space characterization}

The action at each time step is a 4-DoF vector $\mathbf{a}_t = (\dxl, \dyl, \dz, \dpsi)$ expressed in the drone's local body frame. Table \ref{tab:action_stats} reports the marginal statistics of each component across all non-terminal frames ($N = 8{,}504$).

\begin{table}[t]
\renewcommand{\arraystretch}{1.2}
\centering
\caption{Action component statistics over the $N{=}8{,}504$ non-terminal frames.}
\label{tab:action_stats}
\begin{tabular}{lrrrr}
\toprule
\textbf{Component} & \textbf{Mean} & \textbf{Std} & \textbf{Min} & \textbf{Max} \\
\midrule
$\dxl$ (m)   & $+0.864$ & $0.688$ & $-0.558$ & $+5.615$ \\
$\dyl$ (m)   & $-0.012$ & $0.421$ & $-5.124$ & $+4.012$ \\
$\dz$  (m)   & $-0.000$ & $0.013$ & $-0.363$ & $+0.567$ \\
$\dpsi$ (rad)& $+0.096$ & $0.355$ & $-1.254$ & $+2.197$ \\
\bottomrule
\end{tabular}
\end{table}

Four observations characterize the action distribution and motivate later design decisions. Firstly, $\dxl$ is strongly positive on average ($0.86\pm0.69$ m/frame), reflecting the dominant forward-pursuit behavior as the drone closes on or maintains distance from the walking target. Secondly, $\dyl$ is near-zero in mean but with a sizeable standard deviation ($0.42$ m) and a wide range ($[-5.12, +4.01]$ m), capturing frequent and sometimes large lateral correction maneuvers as the target deviates from the drone's heading axis. Thirdly, $\dz$ is tightly concentrated around zero (mean $-0.000$ m, std $0.013$ m), confirming near-constant-altitude flight in the majority of episodes; the small negative minimum ($-0.36$ m) arises from \texttt{height\_change}-tagged episodes in which the operator descends modestly during a long-distance approach. Fourthly, $\dpsi$ exhibits a wide spread (std $0.355$ rad, range $[-1.25, +2.20]$ rad), a direct consequence of the approach/locate-subject episodes that require substantial heading reorientation to acquire the target before pursuit begins. Fig. \ref{fig:actions} shows the full marginal histograms and Fig. \ref{fig:alt_yaw} plots altitude and heading over time for representative episodes.

\begin{widefigure}[tbp]
  \centering
  \includegraphics[width=\linewidth,height=0.72\textheight,keepaspectratio]{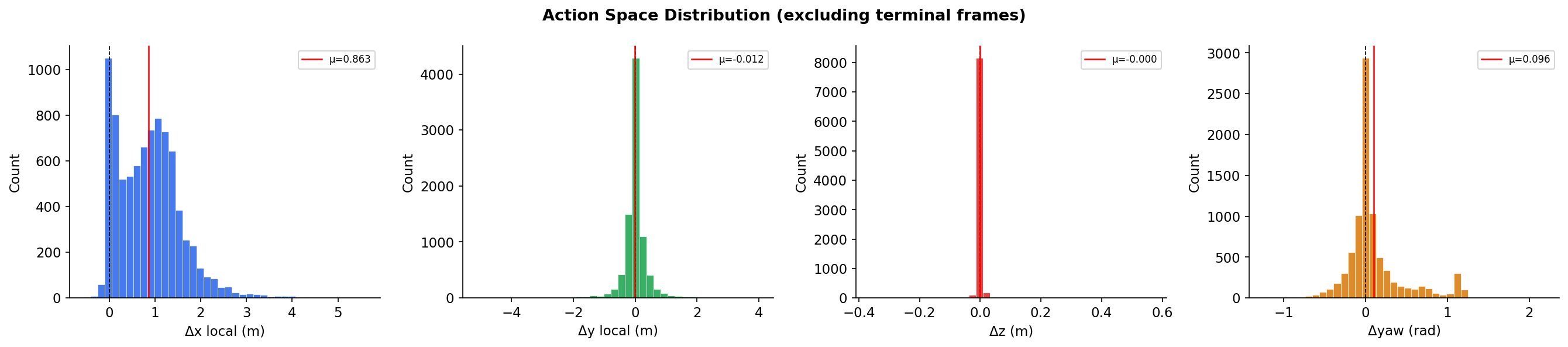}
  \caption{Marginal distributions of the four action components across all non-terminal frames ($N{=}8{,}504$). Dashed red lines indicate per-component means. The strongly right-skewed $\dxl$ reflects persistent forward pursuit; the symmetric $\dyl$ and $\dpsi$ capture bidirectional correction maneuvers.}
  \label{fig:actions}
\end{widefigure}

\begin{widefigure}[tbp]
  \centering
  \includegraphics[width=\linewidth,height=0.72\textheight,keepaspectratio]{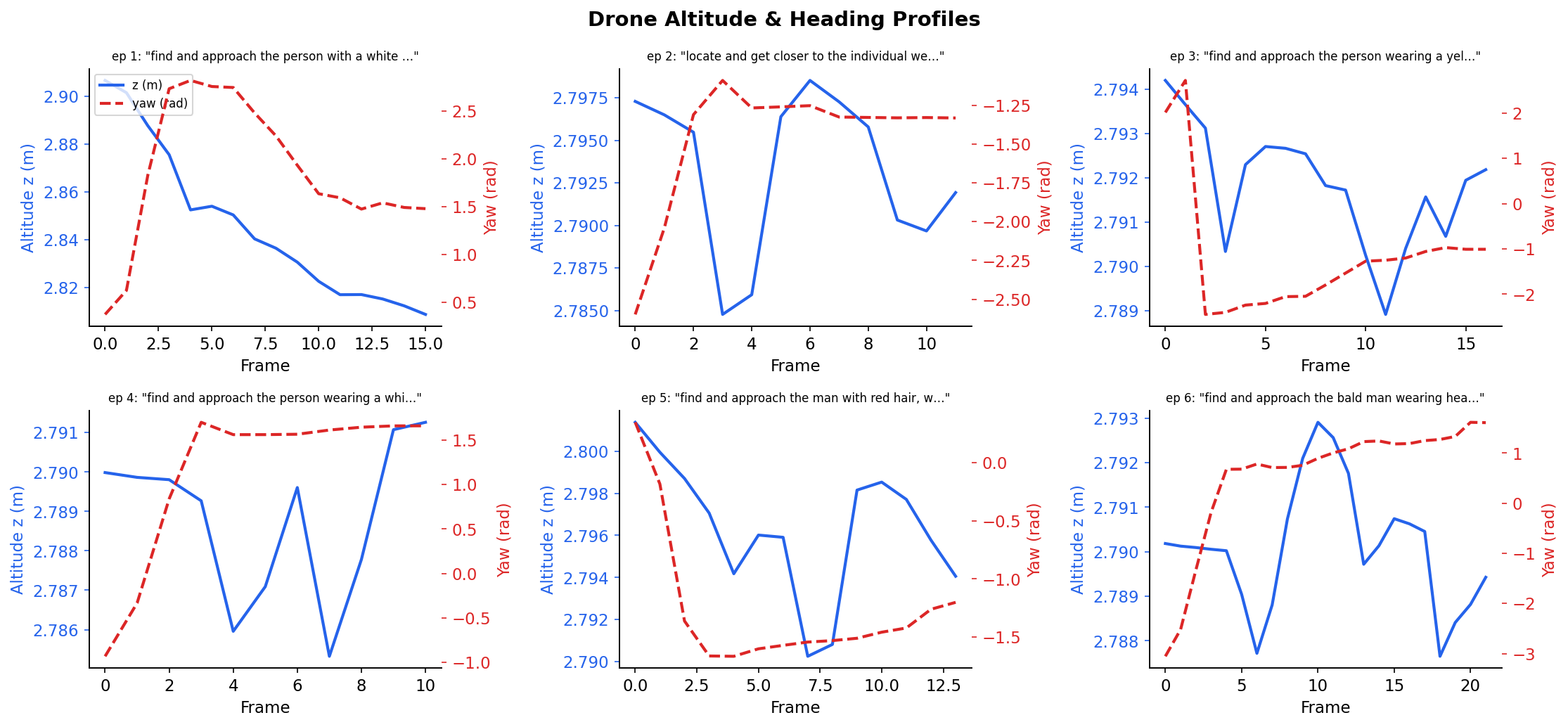}
  \caption{Altitude $z$ (solid blue, left axis) and yaw angle $\psi$ (dashed red, right axis) over time for six episodes. Altitude remains nearly constant in most episodes, while heading changes dynamically as the drone turns to track the moving target.}
  \label{fig:alt_yaw}
\end{widefigure}

\subsection{Linguistic diversity and instruction analysis}

\begin{figure}[tbp]
  \centering
  \includegraphics[width=\linewidth,height=0.72\textheight,keepaspectratio]{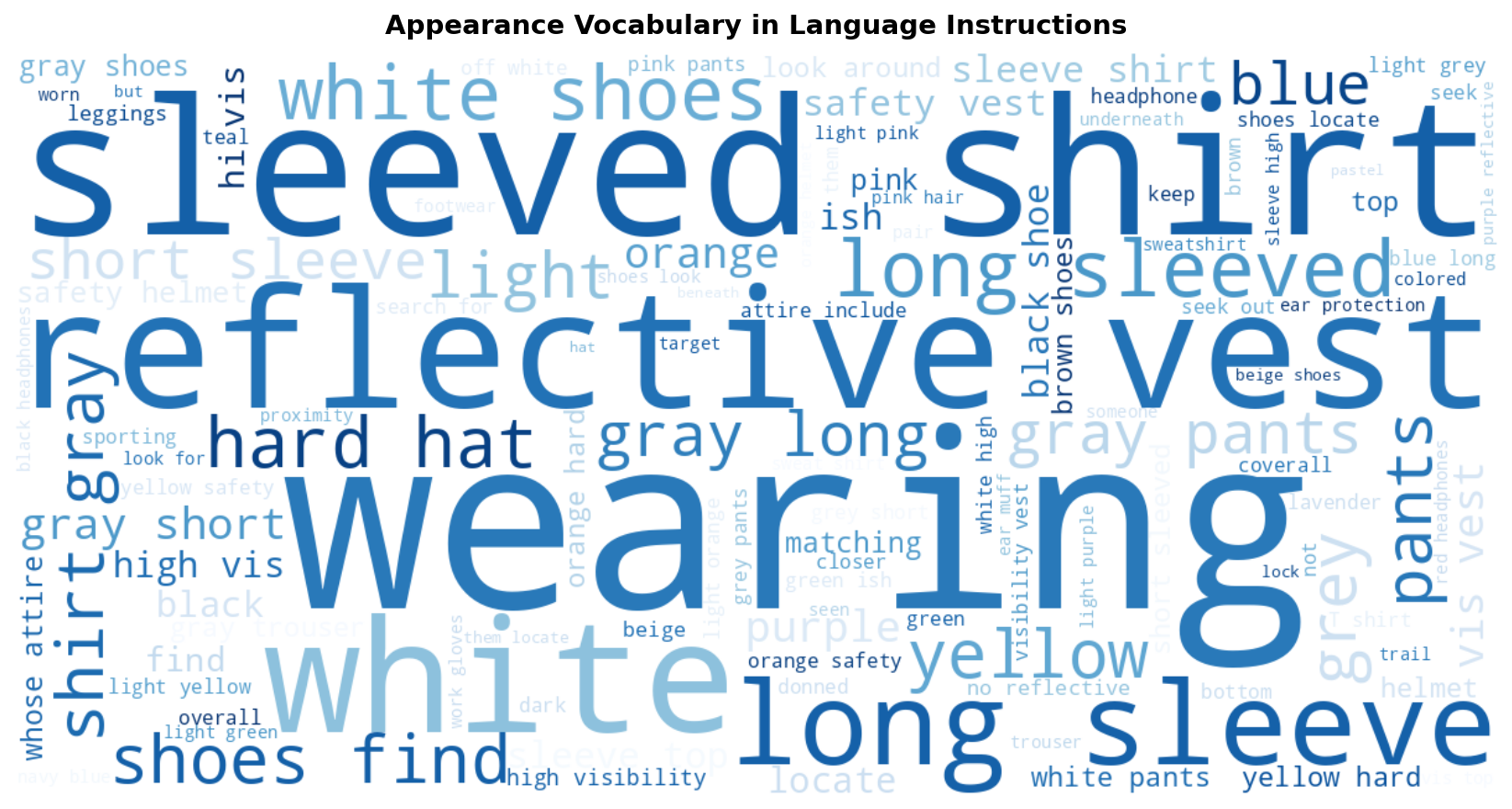}
  \caption{Word cloud of appearance tokens in language instructions (task verbs and stop words removed). Term size reflects relative frequency. PPE vocabulary (\textit{hard hat}, \textit{safety vest}, \textit{headphones}) dominates, alongside rich color and clothing-type diversity.}
  \label{fig:wordcloud}
\end{figure}

Fig. \ref{fig:wordcloud} shows the word cloud of the instruction corpus after removing task verbs and grammatical stop words. The most prominent appearance terms include \textit{safety vest}, \textit{high-visibility}, \textit{reflective}, \textit{hard hat}, \textit{long sleeve}, \textit{overalls}, \textit{headphones} and \textit{boots} lexical items naturally arising from warehouse PPE inventories. The instruction vocabulary also exhibits rich color diversity (\textit{orange}, \textit{purple}, \textit{terracotta}, \textit{teal}, \textit{lavender}, \textit{moss green}) and fine-grained compositional descriptions (\textit{short-sleeve beneath a safety shirt}, \textit{long sleeves beneath a safety shirt}) paraphrase patterns that are known to break naive token-matching grounding systems.

The corpus spans $15$ task-verb families dominated by \emph{approach} (the approach/locate-subject regime) and \emph{follow} (continuous pursuit), with thirteen additional verb families including \emph{accompany}, \emph{pursue}, \emph{go after}, \emph{move closer}, \emph{move towards}, \emph{fly after}, \emph{trace}, \emph{tail}, \emph{stay close} and \emph{move along} providing paraphrase diversity (Fig. \ref{fig:overview}c). This design prevents models from ignoring the task-verb token, as different verbs may imply subtly different behavioral profiles (\emph{e.g.}, \emph{approach} implying active closure vs.\ \emph{accompany} implying lateral co-navigation). The dataset thus constitutes a natural testbed for instruction paraphrase generalization, a key challenge identified in recent VLN and VLA literature \cite{liu2023aerialvln,wang2024uav}; Table~\ref{tab:verb_breakdown} (\ref{app:robustness}) reports per-verb-family error and confirms that the model ranking is preserved across all linguistic conditions.

\subsection{Challenge tags and difficulty axes}

A subset of episodes carries explicit difficulty annotations that mark independent generalization axes. The most common is \textit{find}, which marks the approach/locate-subject regime in which the drone must first acquire a specified worker before initiating pursuit. The remainder span \textit{close} (the drone operates at close range to the target), \textit{far} (the target reaches a distance ${>}10$ m), \textit{height\_change} (the drone changes altitude by ${>}0.5$ m during the episode), \textit{obstructed} (the target passes fully behind warehouse infrastructure for at least one frame) and \textit{extended} (an unusually long trajectory). These tags support stratified evaluation along independent generalization axes. Fig. \ref{fig:frames} shows sample frames from representative episodes.

\begin{widefigure}[tbp]
  \centering
  \includegraphics[width=\linewidth,height=0.72\textheight,keepaspectratio]{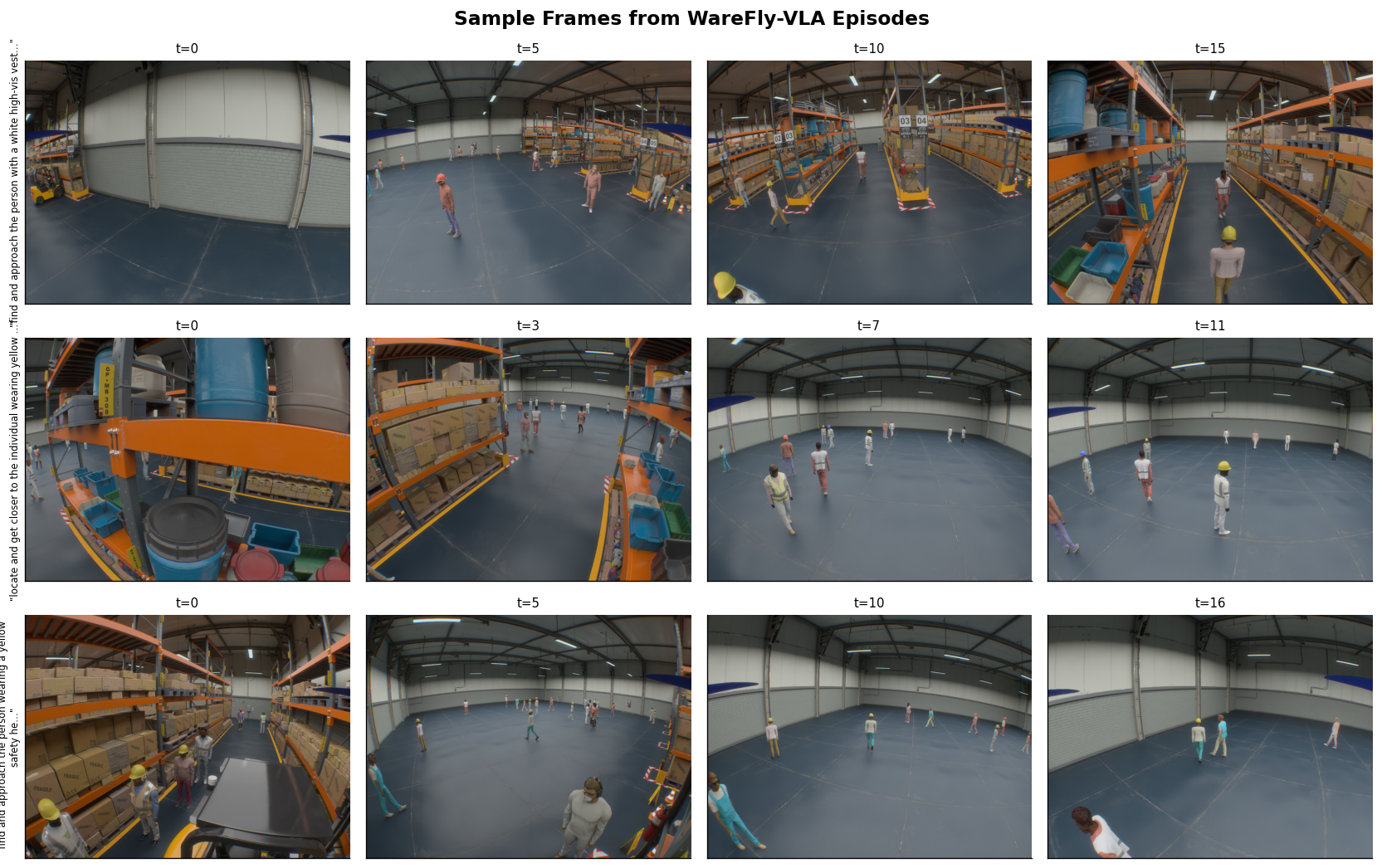}
  \caption{Sample frames from three episodes at four evenly-spaced time steps (left to right: $t{=}0, t_1, t_2, t_\text{end}$). Row labels show the corresponding language instruction (truncated). The warehouse background, diverse worker appearances and varying drone--target distances illustrate the visual complexity of the dataset.}
  \label{fig:frames}
\end{widefigure}

\subsection{Comparison with related datasets}

Table \ref{tab:comparison} positions WareFly-VLA against the most closely related existing datasets across the dimensions of platform, language supervision, action representation, data source and industrial relevance. WareFly-VLA is the only dataset combining a UAV platform, natural-language appearance-based target specification, \emph{continuous low-level action supervision} and an \emph{industrial/warehouse environment}; we obtain these properties through photorealistic simulation rather than physical capture.

\begin{table}[t]
\renewcommand{\arraystretch}{1.0}
\renewcommand\theadfont{\bfseries}
\centering
\caption{Comparison with related datasets.
  \checkmark{}~=~yes, $\times$~=~no, --~=~not applicable.
  ``Cont.\ actions'' indicates whether continuous low-level action labels
  (not high-level discrete commands or waypoints) are provided;
  ``Scr.'' indicates real-world capture vs.\ simulation.}
\label{tab:comparison}
\begin{tabular}{lccccc}
\toprule
\textbf{Dataset} & \thead{Plat-\\form} & \thead{Lang.\\ instr.}
  & \thead{Cont.\\ acts.} & \textbf{Src.}
  & \textbf{Episodes} \\
\midrule
UAV123 \cite{mueller2016uav}                & UAV    & $\times$     & \checkmark   & Real & $\sim$150 \\
AerialVLN \cite{liu2023aerialvln}           & UAV    & \checkmark   & $\times$     & Sim  & 8,446 \\
AirNav \cite{deshpande2026airnav}           & UAV    & \checkmark   & $\times$     & Real & 35,000 \\
UAV-VLA \cite{sautenkov2025uavvla}          & UAV    & \checkmark   & $\times$     & Real & 100k \\
RaceVLA \cite{serpiva2025racevla}           & UAV    & \checkmark   & \checkmark   & Real & $\sim$200 \\
CognitiveDrone \cite{cogdrone2025}          & UAV    & \checkmark   & \checkmark   & Sim  & 8,000 \\
BridgeData V2 \cite{walke2023bridgedata}    & Arm    & \checkmark   & \checkmark   & Real & 60,096 \\
DROID \cite{khazatsky2024droid}             & Arm    & \checkmark   & \checkmark   & Real & 76,000 \\
Open~X-Embodiment \cite{padalkar2023rtx}    & Multi  & \checkmark   & \checkmark   & Real & $1$M+ \\
\midrule
\textbf{WareFly-VLA (ours)}                 & \textbf{UAV} & \checkmark
  & \checkmark   & Sim  & \textbf{507} \\
\bottomrule
\end{tabular}
\end{table}

We emphasize that the relatively modest episode count compared to manipulating datasets is not an oversight: each episode requires manual teleoperation paired with a hand-authored, uniquely identifying worker appearance and the dataset's primary value lies in its \emph{task fidelity} and \emph{linguistic diversity} photorealistic warehouse imagery, varied PPE configurations and human-flown target-following trajectories rather than in raw scale. The benchmark experiments in Sections \ref{sec:experiments} and \ref{sec:results} demonstrate that, despite this modest scale, the dataset already discriminates sharply between modern VLA architectures.

\section{Benchmark protocol}
\label{sec:experiments}

\subsection{Benchmark models}

A primary objective of WareFly-VLA is to provide a standardized benchmark for evaluating Vision--Language--Action learning in aerial environments. To this end, four representative open-source VLA architectures are selected: OpenVLA-7B \cite{kim2024openvla}, GR00T N1.7 \cite{nvidia2025groot}, $\pi_0$/OpenPI \cite{black2024pi0} and SmolVLA \cite{shukor2025smolvla}. These models collectively span the three dominant action-generation paradigms currently used in VLA systems, namely discrete action tokenization, diffusion-based action generation and flow matching. They further cover a broad range of parameter scales and embodiment assumptions, providing a diverse testbed for studying aerial transfer and language-conditioned UAV control. Table \ref{tab:baselines} summarizes the benchmarked architecture used for fine-tuning.

Because the same trajectories can be re-exported at an arbitrary control rate, we also evaluate at $10$ FPS the deployment rate of a reactive body-frame flight loop in addition to the $1$ FPS rate at which contemporary VLAs publish results. To keep the cross-rate comparison interpretable under our compute budget, at $10$ FPS we re-train only the \emph{accuracy frontier} $\pi_0$ and SmolVLA, the two models that beat the constant-mean baseline at $1$ FPS at matched optimization budgets ($15{,}000$ steps each). OpenVLA, whose epoch-based recipe would exceed $50$ h on the $10\times$-larger corpus and GR00T, which diverges at $1$ FPS (Section \ref{sec:results}), are excluded at $10$ FPS as uninformative.

\begin{table*}[t]
\centering
\caption{Representative Vision--Language--Action architectures benchmarked on WareFly-VLA.}
\label{tab:baselines}

\small
\begin{tabularx}{\textwidth}{l c X X c}
\toprule
\textbf{Model} &
\textbf{Parameters} &
\textbf{Action Representation} &
\textbf{Vision--Language Backbone} &
\textbf{Adaptation} \\
\midrule

OpenVLA-7B &
\makecell[c]{7.6B total\\49.3M trainable} &
\makecell[l]{Autoregressive policy\\(256 bins per dimension)} &
\makecell[l]{Prismatic\\SigLIP + DINOv2\\Llama-2 7B} &
LoRA \\

GR00T N1.7 &
\makecell[c]{3.1B total\\1.62B trainable} &
\makecell[l]{Diffusion policy\\Horizon = 4} &
\makecell[l]{Cosmos-Reason2-2B} &
Fine-tuning \\

$\pi_0$/OpenPI &
\makecell[c]{3.3B total\\145M trainable} &
\makecell[l]{Flow matching\\Horizon = 4} &
\makecell[l]{PaliGemma-3B} &
LoRA \\

SmolVLA &
\makecell[c]{450M total\\450M trainable} &
\makecell[l]{Flow matching\\Horizon = 4} &
\makecell[l]{SmolVLM-2\\+ Action Expert} &
Full fine-tuning \\

\bottomrule
\end{tabularx}
\end{table*}

\subsection{Data splits}
\label{sec:splits}

A single canonical episode-level split is adopted throughout the benchmark. Episodes are partitioned using a fixed random seed and a validation ratio of 15\%, yielding 431 training episodes and 76 validation episodes (7{,}276 train / 1{,}228 val non-terminal transitions at $1$ FPS; 73{,}979 / 13{,}371 transitions at $10$ FPS) in the current dataset release. The $431$/$76$ episode partition is shared verbatim by every model at both temporal resolutions, so no validation episode enters any training set at either rate. Splitting is performed at the trajectory level rather than the frame level to eliminate temporal leakage between training and evaluation samples.

\subsection{Training protocol}

All models are initialized from publicly released checkpoints and fine-tuned exclusively on WareFly-VLA training data. Whenever possible, official implementations and recommended training procedures provided by the original authors are adopted to minimize implementation-specific bias.

Input observations consist of a single RGB image and its associated language instruction. Images are resized to the native input resolution expected by each architecture. For models whose native action spaces differ from the four-dimensional UAV control representation used in WareFly-VLA, compatibility mappings are applied while preserving the semantic meaning of the original control commands.

No test-time augmentation is employed. Unless otherwise stated, all reported results correspond to a single forward pass per validation sample.

\subsection{Evaluation metrics}

For each action dimension $d \in \{\Delta x,\Delta y,\Delta z,\Delta\psi\}$, Mean Absolute Error (MAE) and Pearson correlation coefficient are described by

\begin{align}
\text{MAE}_d &= \frac{1}{N}\sum_{t=1}^{N}\left|\hat{a}_{t,d}-a_{t,d}\right|,\\
r_d &= \frac{\sum_t(\hat{a}_{t,d}-\bar{\hat{a}}_d)(a_{t,d}-\bar{a}_d)}{\sqrt{\sum_t(\hat{a}_{t,d}-\bar{\hat{a}}_d)^2}\sqrt{\sum_t(a_{t,d}-\bar{a}_d)^2}}.
\end{align}

MAE measures action prediction fidelity, whereas Pearson correlation evaluates how well a policy reproduces the temporal structure of expert behavior. These metrics provide complementary perspectives on policy quality, as low prediction error may be achieved by regressing toward the mean action, while high correlation does not guarantee accurate action magnitudes.

In addition, a constant-mean baseline is reported for each action dimension. This baseline predicts the mean action observed in the training set and provides a lower bound for meaningful policy learning.

\section{Results and analysis}
\label{sec:results}

\subsection{Main results}

\begin{figure}[tbp]
  \centering
  \includegraphics[width=\linewidth,height=0.72\textheight,keepaspectratio]{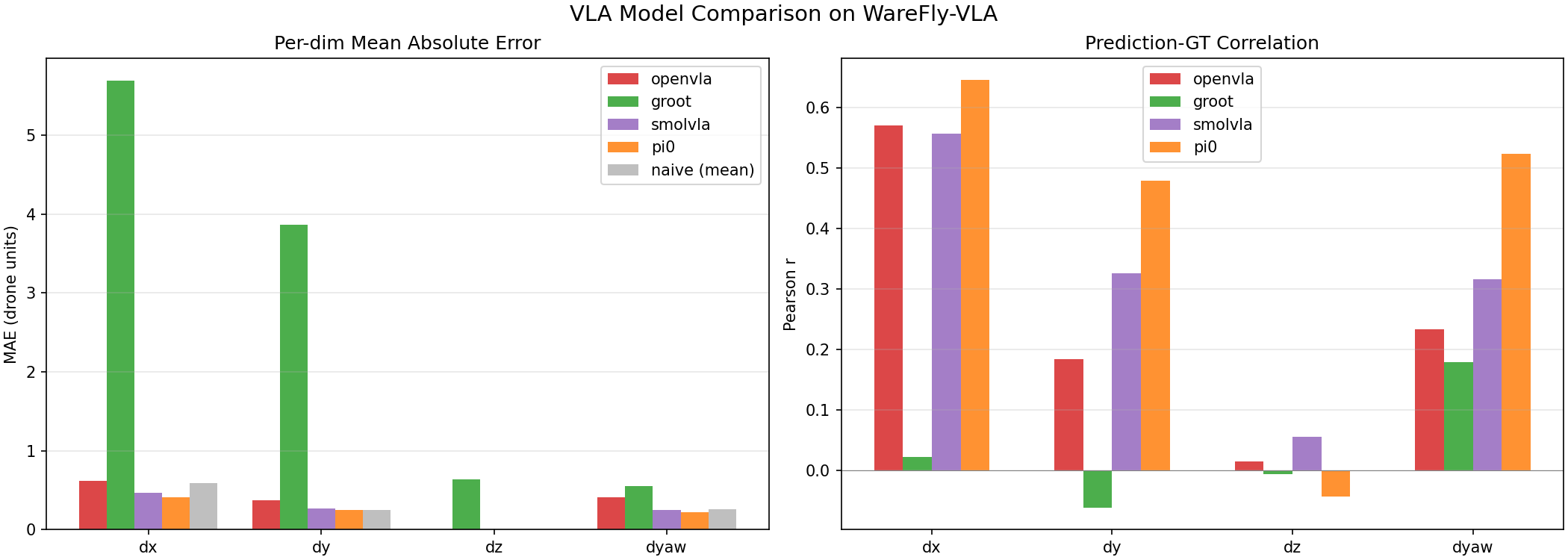}
  \caption{Cross-model comparison on the WareFly-VLA held-out validation split (76 identical episodes for all models). \textbf{Left:} Per-dimension MAE (lower is better). \textbf{Right:} Pearson correlation (higher is better). $\pi_0$ achieves the best overall performance, followed by SmolVLA, while OpenVLA shows limited generalization and GR00T fails to produce meaningful predictions. No model surpasses the naive baseline on $\dyl$ or $\dz$, highlighting the difficulty of lateral and vertical action prediction.}
  \label{fig:model_comparison}
\end{figure}

Table \ref{tab:main_results} reports per-dimension MAE and correlation of each model on the held-out validation episodes, together with the naive constant-mean reference. Figure \ref{fig:model_comparison} visualizes these numbers across all four dimensions; the $y$-axis range is preserved so that GR00T's outsized errors remain visible.

\begin{table}[t]
\renewcommand{\arraystretch}{1.25}
\centering
\caption{Per-step prediction performance on the WareFly-VLA held-out validation split. Results are reported as Mean Absolute Error (MAE) and Pearson correlation ($r$). All models are evaluated on the same 76 held-out episodes ($N=1,228$ frames) under the canonical leakage-free split.}
\label{tab:main_results}
\begin{tabular}{lcccc}
\toprule
\textbf{Model} & $\dxl$ (m) & $\dyl$ (m) & $\dz$ (m) & $\dpsi$ (rad) \\
\midrule
\multicolumn{5}{c}{\emph{MAE $\downarrow$}} \\
\midrule
Naive (constant-mean) & 0.590 & 0.242 & 0.003 & 0.248 \\
\midrule
OpenVLA-7B~\cite{kim2024openvla} & \underline{0.620} & \underline{0.368} & \underline{0.005} & \underline{0.409} \\
GR00T~N1.7~\cite{nvidia2025groot}$^{\dagger}$ & \underline{5.728} & \underline{3.867} & \underline{0.637} & \underline{0.555} \\
$\pi_0$/OpenPI~\cite{black2024pi0} & \textbf{0.413} & \textbf{\underline{0.244}} & \textbf{\underline{0.003}} & \textbf{0.220} \\
SmolVLA~\cite{shukor2025smolvla} & 0.468 & \underline{0.264} & \underline{0.003} & 0.243 \\
\midrule
\multicolumn{5}{c}{\emph{Pearson $r \uparrow$}} \\
\midrule
OpenVLA-7B & +0.57 & +0.18 & +0.01 & +0.23 \\
GR00T~N1.7$^{\dagger}$ & +0.01 & -0.06 & -0.01 & +0.18 \\
$\pi_0$/OpenPI & \textbf{+0.65} & \textbf{+0.48} & -0.04 & \textbf{+0.52} \\
SmolVLA & +0.56 & +0.33 & \textbf{+0.06} & +0.32 \\
\bottomrule
\end{tabular}
\end{table}

The headline result, once train/validation leakage is removed (Section \ref{sec:splits}), is that \emph{WareFly-VLA is far from solved}. At $1$ FPS the strongest model, $\pi_0$, attains only $r{=}0.65$ on the forward displacement $\dxl$, $r{=}0.52$ on heading $\dpsi$ and $r{=}0.48$ on the lateral channel $\dyl$, while the vertical channel $\dz$ is at chance ($|r|{\le}0.06$ for every model). \emph{No} model beats the naive constant-mean baseline in MAE on $\dyl$ or $\dz$. The flow-matching models form the top tier: $\pi_0$ (3.3B) leads SmolVLA (450M) on every dimension ($\dxl$ $r$ $0.65$ vs.\ $0.56$; $\dyl$ $r$ $0.48$ vs.\ $0.33$; $\dpsi$ $r$ $0.52$ vs.\ $0.32$) and both beat naive in MAE on $\dxl$ and $\dpsi$. OpenVLA recovers forward \emph{direction} reasonably ($\dxl$ $r{=}0.57$) but is \emph{worse than naive on every MAE channel}, because it systematically overshoots magnitude on the in-plane channels and collapses the low-amplitude $\dz$ and $\dpsi$ commands onto quantization-bin centers. The most important qualitative finding is shared across all models: only the forward channel $\dxl$ is reliably learnable from a single frame, while lateral, vertical and (for all but $\pi_0$ and SmolVLA) heading commands are not these channels demand temporal context that a per-frame policy cannot access (Section \ref{sec:discussion}).

\begin{figure}[tbp]
\centering
\includegraphics[width=\linewidth,height=0.72\textheight,keepaspectratio]{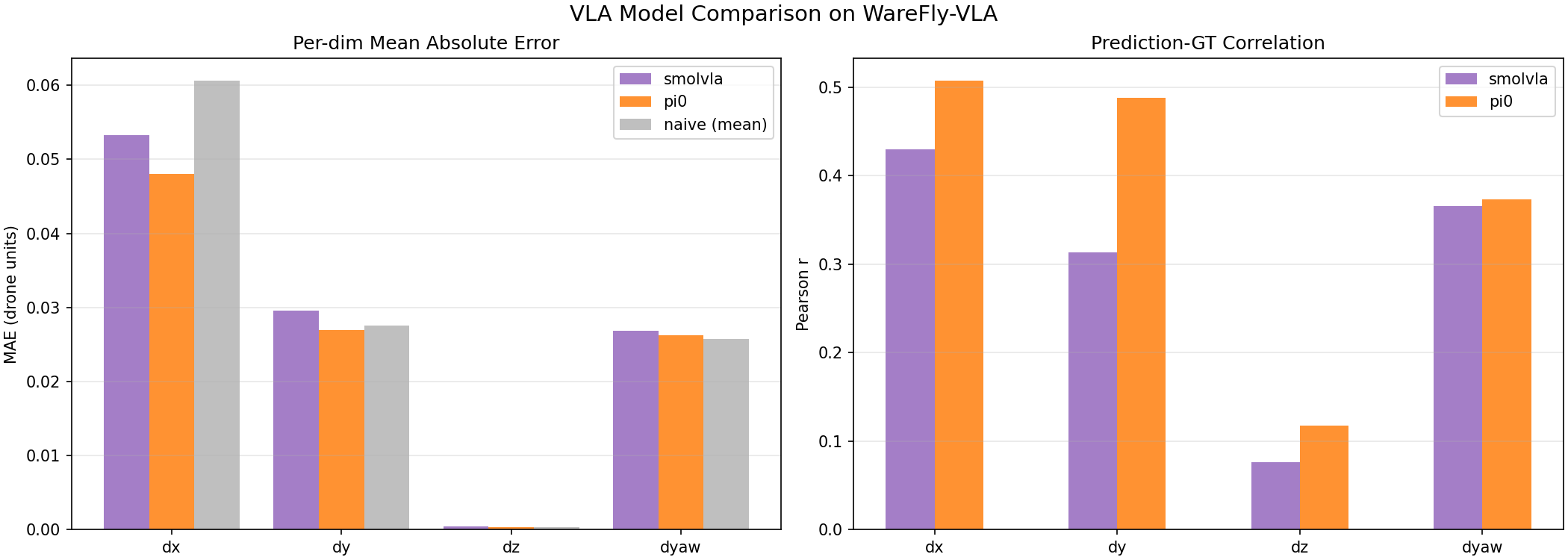}
\caption{Cross-model comparison at the $10$ FPS deployment-rate setting. \textbf{Left:} per-dimension MAE learned bars sit nearly flush with the naive baseline on $\dyl$/$\dz$/$\dpsi$, the consequence of $\sim$$10\times$-smaller per-frame actions. \textbf{Right:} Pearson $r$ preserves the $1$ FPS structure ($\pi_0\!>\!$SmolVLA; forward most learnable, vertical least). The contrast motivates reporting correlation alongside MAE at high control rates.}
\label{fig:model_comparison_10fps}
\end{figure}

The same per-step protocol is also applied at the $10$ FPS deployment-rate setting for the accuracy frontier ($\pi_0$, SmolVLA). Table \ref{tab:main_results_10fps} reports the corresponding per-dimension MAE and Pearson $r$ on the identical $76$ held-out episodes ($N{=}13{,}371$ transitions). Two observations stand out. \emph{First, the architecture ordering is preserved}: $\pi_0 \!>\!$ SmolVLA $\!>\!$ naive on $\dxl$ in MAE and $\pi_0$ leads SmolVLA in correlation on the forward and lateral channels ($\dxl$ $r$ $0.51$ vs.\ $0.43$; $\dyl$ $r$ $0.49$ vs.\ $0.31$). \emph{Second, the per-step MAE margin over the baseline collapses}: because each action spans one inter-frame interval, the $10$ FPS per-frame displacements are $\sim$$10\times$ smaller and the naive constant-mean baseline is correspondingly tighter, so only $\dxl$ is clearly beaten in MAE by both models; on $\dyl$ only $\pi_0$ edges the baseline and on $\dpsi$/$\dz$ \emph{neither} model beats naive in MAE despite $\pi_0$ retaining a positive heading correlation ($r{=}0.37$). To make the comparison dimensionless and baseline-relative across rates we further report the skill score $s_d = 1 - \textsc{MAE}_d / \textsc{MAE}_d^{\text{naive}}$ in the discussion (Section \ref{sec:discussion}); $s_d{>}0$ beats the same-rate baseline. Fig. \ref{fig:model_comparison_10fps} visualizes the cross-model comparison at $10$ FPS: the correlation panel preserves the $1$ FPS structure, while the MAE panel sits nearly flush with the baseline on the low-amplitude channels.

\begin{table}[t]
\renewcommand{\arraystretch}{1.25}
\centering
\caption{Per-step prediction accuracy at $10$\,FPS on the validation split ($76$ episodes, $N{=}13{,}371$ transitions). Values report MAE and Pearson $r$ with $95\%$ bootstrap confidence-interval half-widths.}
\label{tab:main_results_10fps}
\setlength{\tabcolsep}{3pt}
\small
\begin{tabular}{lcccc}
\toprule
\textbf{Model} & $\dxl$ (m) & $\dyl$ (m) & $\dz$ (m) & $\dpsi$ (rad) \\
\midrule \multicolumn{5}{c}{\emph{MAE $\downarrow$ ($\times10^{-2}$, except $\dz$)}} \\ \midrule
Naive    & $6.04 {\pm} 0.09$ & $2.74 {\pm} 0.07$ & $0.028 {\pm} 0.001$ & $2.58 {\pm} 0.05$ \\
$\pi_0$  & \textbf{4.80} {${\pm}$} 0.09 & \textbf{2.70} {${\pm}$} 0.06 & \underline{0.037} & \underline{2.62} {${\pm}$} 0.06 \\
SmolVLA  & 5.33 {${\pm}$} 0.09 & \underline{2.95} {${\pm}$} 0.06 & \underline{0.038} & \underline{2.68} {${\pm}$} 0.06 \\
\midrule \multicolumn{5}{c}{\emph{Pearson $r \uparrow$}} \\ \midrule
$\pi_0$  & \textbf{$+0.51$} & \textbf{$+0.49$} & \textbf{$+0.12$} & \textbf{$+0.37$} \\
SmolVLA  & $+0.43$ & $+0.31$ & $+0.08$ & $+0.37$ \\
\bottomrule
\end{tabular}
\end{table}

\subsection{Generalization gap}
\label{sec:gen-gap}

The gap between training-set and held-out accuracy is the single most informative statistic in this benchmark and it is large for every model. Table~\ref{tab:train_val_gap} (\ref{app:robustness}) quantifies it: on $\dxl$, SmolVLA's MAE rises from $0.083$ (train) to $0.468$ (val), a $5.6\times$ degradation; $\pi_0$ rises from $0.150$ to $0.413$ ($2.8\times$); OpenVLA from $0.472$ to $0.620$ ($1.3\times$). Two patterns emerge. First, the \emph{smaller} flow-matching model overfits the hardest: SmolVLA achieves the best training fit of any model yet generalizes worse than the $7\times$-larger $\pi_0$, so its capacity is spent memorizing the $431$ training episodes rather than acquiring a transferable visuomotor mapping. Second, OpenVLA exhibits the smallest \emph{relative} gap only because it underfits even the training data, a direct consequence of quantizing a continuous, heavy-tailed action distribution into $256$ bins. This train/val gap is precisely the quantity that train-on-everything evaluation protocols hide; under the leakage-free split it becomes the defining challenge of the benchmark.

The same overfitting signature persists at $10$ FPS and again favours the larger model: $\pi_0$'s $\dxl$ MAE rises from $0.014$ (train) to $0.048$ (val), a $+0.034$ gap, versus SmolVLA's $0.013\to0.053$ ($+0.040$). At both rates SmolVLA attains the lower training error yet the larger held-out gap, confirming that the capacity-versus-overfitting dissociation is a property of the architectures rather than of the chosen sampling rate.

\subsection{Per-sample prediction quality}

Fig. \ref{fig:scatter_pi0} resolves the per-sample prediction quality of the leading model ($\pi_0$) into a scatter plot for each action dimension on the held-out split. The $\dxl$ panel shows a clear positive trend along the identity diagonal but with substantial residual scatter the model captures the \emph{direction} of forward motion yet compresses its dynamic range, under-predicting the largest pursuit displacements. The $\dyl$ and $\dz$ panels collapse toward a horizontal band near the training mean, the visual signature of a policy that has defaulted to the prior because the lateral and vertical commands are not predictable from a single frame. This contrasts sharply with the corresponding train-split scatter (not shown), where all four dimensions hug the diagonal a direct visualization of the generalization gap quantified in Section \ref{sec:gen-gap}.

\begin{widefigure}[tbp]
  \centering
  \includegraphics[width=\linewidth,height=0.72\textheight,keepaspectratio]{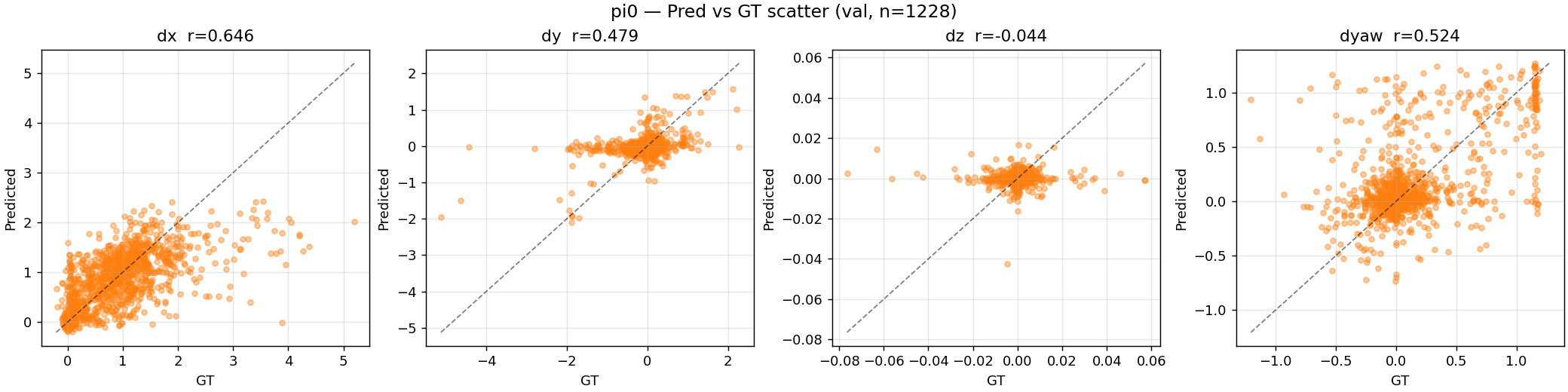}
  \caption{Per-dimension prediction-vs.-ground-truth scatter for $\pi_0$ (the strongest model) on the held-out validation split. The dashed diagonal denotes the identity line; $\dxl$ shows a positive but range-compressed trend, while $\dyl$ and $\dz$ collapse toward the prior mean the per-sample signature of the generalization gap.}
  \label{fig:scatter_pi0}
\end{widefigure}

\subsection{Qualitative analysis}

\begin{figure}[tbp]
  \centering
  \includegraphics[width=\linewidth,height=0.72\textheight,keepaspectratio]{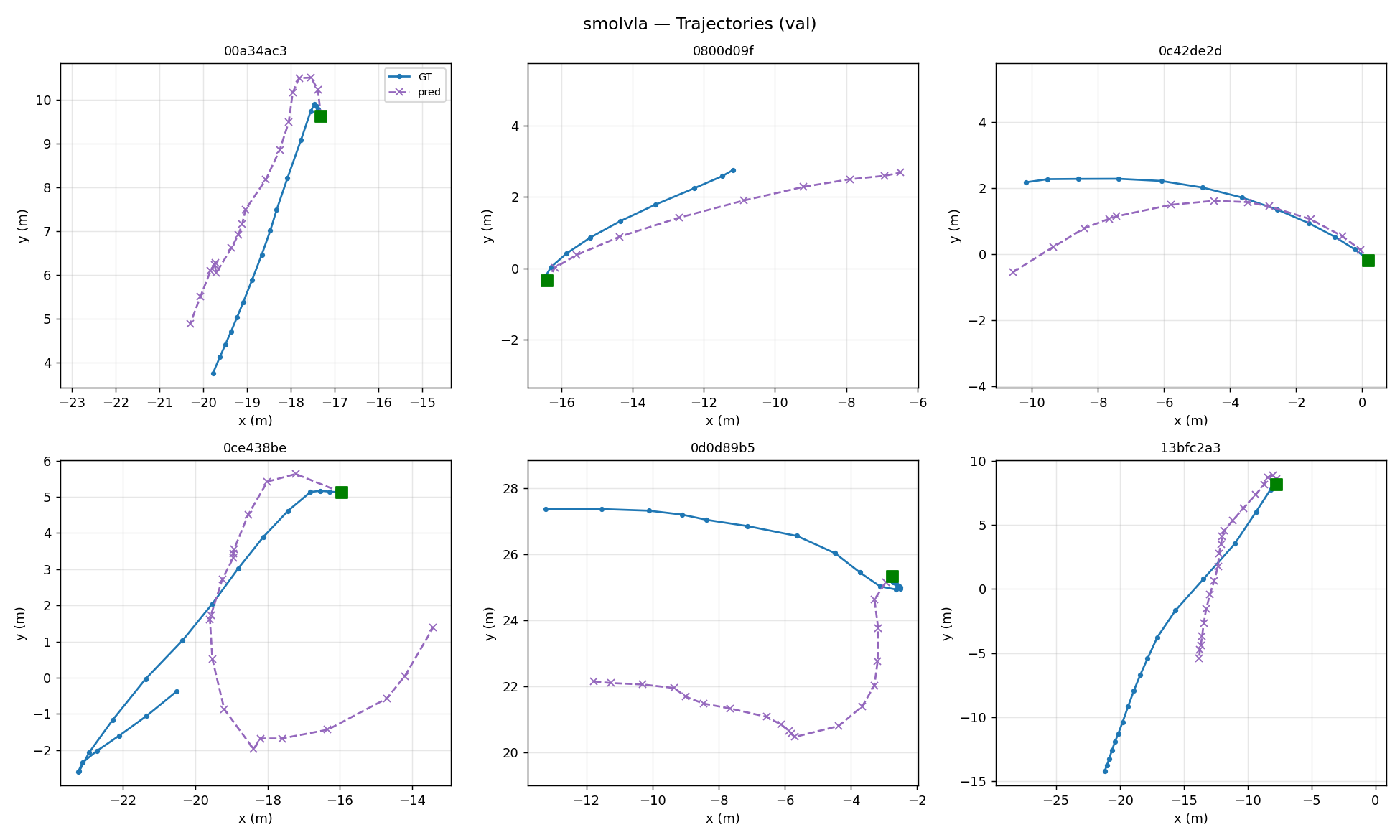}
  \caption{Open-loop trajectory rollouts of SmolVLA predictions vs.\ expert trajectories on six held-out validation episodes. Solid blue is the expert path; dashed purple is the integrated prediction starting from the true initial pose; green squares mark start positions. Rollouts follow the gross forward progression but accumulate drift on curved segments where lateral/heading corrections are mi-predicted.}
  \label{fig:traj_smolvla}
\end{figure}

Fig. \ref{fig:qualitative} visualizes SmolVLA's predicted actions overlaid on representative held-out frames. For each frame we draw the ground-truth in-plane displacement $(\dxl, \dyl)$ as a solid blue arrow and the SmolVLA prediction as a dashed purple arrow originating at the same image-plane anchor; text overlay reports the per-dimension ground-truth, prediction and absolute error. The predicted arrows are consistently oriented forward, tracking the dominant $\dxl$ component but they systematically \emph{under-reach} the ground-truth arrow on large displacements and frequently mis-estimate the lateral $\dyl$ component the per-frame manifestation of the near-zero held-out lateral correlation in Table \ref{tab:main_results}. We further probe \emph{rolled-out trajectory fidelity} by integrating each predicted action sequence forward in time from the true initial pose of each episode (Fig. \ref{fig:traj_smolvla}). On held-out episodes the integrated rollouts capture the gross forward progression of the expert path but accumulate visible drift on curved segments, where the unpredicted lateral and heading corrections compound over time a direct consequence of the single-frame observation model.

\begin{widefigure}[tbp]
  \centering
  \includegraphics[width=\linewidth,height=0.72\textheight,keepaspectratio]{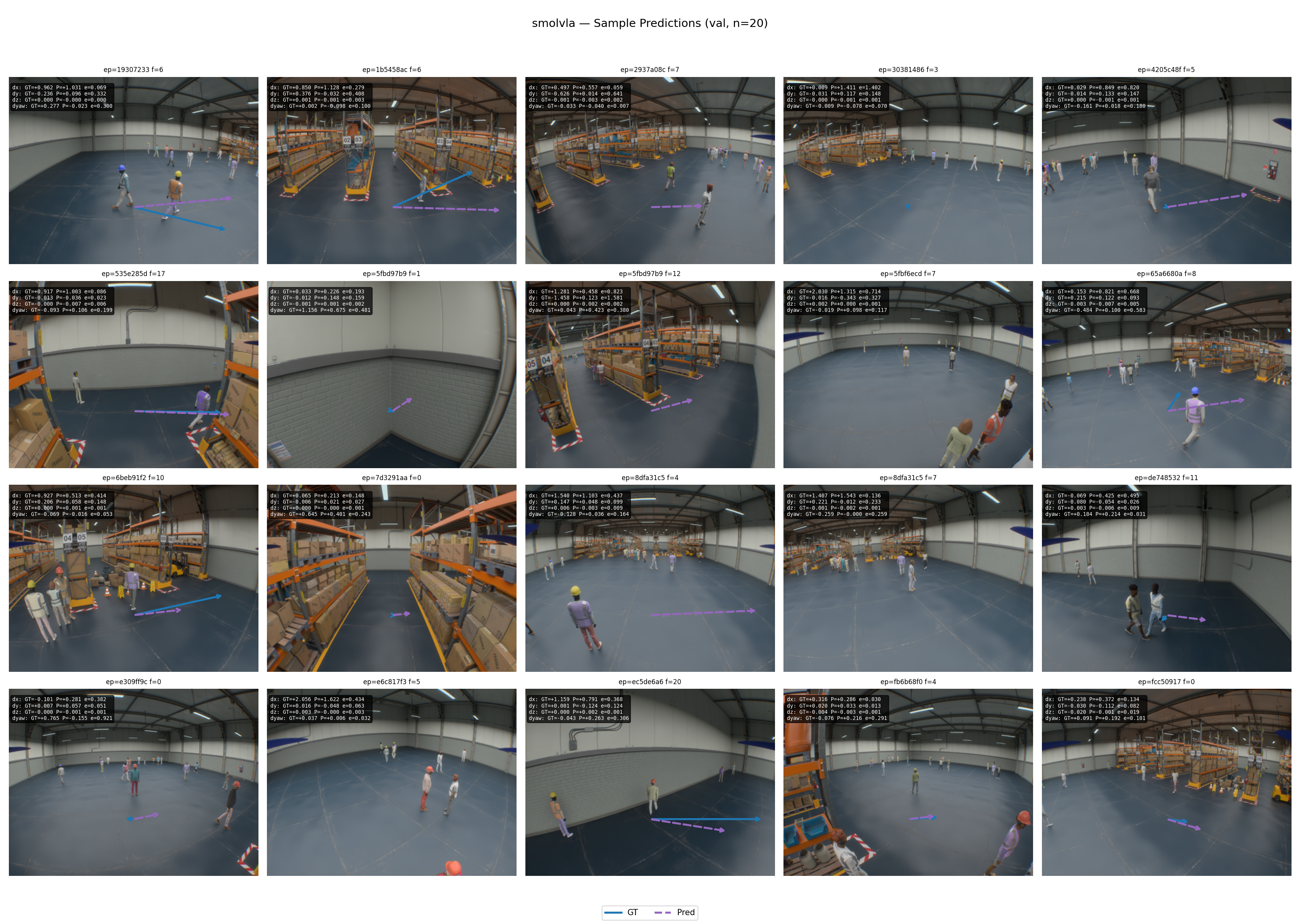}
  \caption{Qualitative SmolVLA predictions on twenty randomly-sampled
    \emph{held-out} frames.
    Solid blue arrows depict the ground-truth in-plane displacement;
    dashed purple arrows depict the predicted displacement.
    Predictions track the forward direction but under-reach on large
    displacements and miss lateral corrections, illustrating the
    generalization gap on truly unseen episodes.}
  \label{fig:qualitative}
\end{widefigure}

\section{Deployment-oriented analysis}
\label{sec:deployment}

\subsection{Cost--accuracy frontier}

\begin{figure}[tbp]
  \centering
  \includegraphics[width=\linewidth,height=0.72\textheight,keepaspectratio]{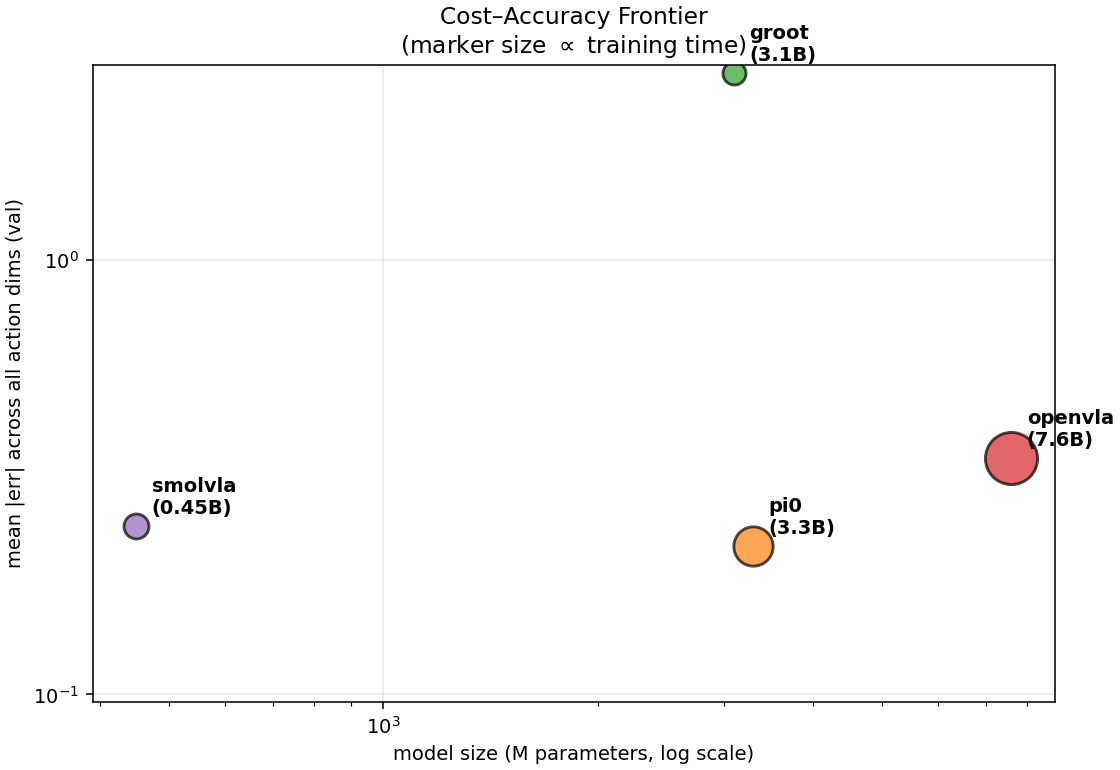}
    \caption{Cost--accuracy frontier. $x$-axis: model size (log scale); $y$-axis: mean validation MAE (symmetric log); marker area $\propto$ fine-tuning wall-clock time. SmolVLA and $\pi_0$ jointly define the Pareto frontier (cheapest vs.\ most accurate); OpenVLA is dominated and GR00T fails.}
  \label{fig:cost_accuracy}
\end{figure}

Fig. \ref{fig:cost_accuracy} positions each model on a two-dimensional chart of parameter count (log scale) and mean validation MAE (symmetric log), with marker area encoding fine-tuning wall-clock time on a single H100~$80$GB GPU. The Pareto frontier is defined by exactly two models: SmolVLA (smallest at $450$M and fastest to fine-tune) and $\pi_0$ (most accurate on the held-out split). The choice between them is a genuine cost--accuracy trade-off rather than a domination: SmolVLA is $7\times$ smaller and trains in a fraction of the time, while $\pi_0$ buys the best held-out correlation on the two learnable dimensions at $7\times$ the parameter count. OpenVLA-7B is dominated $17\times$ larger than SmolVLA, slower to fine-tune and less accurate on every dimension and GR00T, comparable in size to $\pi_0$, fails catastrophically. The actionable recommendation is therefore nuanced: under a tight compute budget SmolVLA is the rational default, but when held-out fidelity matters $\pi_0$ is preferable; critically, \emph{neither} reaches deployable accuracy on this benchmark, which remains open.

\subsection{Inference latency and control rate}
\label{sec:latency}

Per-step accuracy is silent on what ultimately decides deployability on a drone: whether a policy can close its control loop in real time. Table \ref{tab:latency} reports single-frame inference latency on one H100 ($10$ warm-up, $50$ timed calls). At $1$ FPS, SmolVLA is fastest ($36.5$ ms, $27.4$ Hz) and $\pi_0$ sustains $14.9$ Hz, while OpenVLA-7B manages only $2.7$ Hz its autoregressive decoder cannot keep pace with a reactive loop and GR00T is fast but produces unusable predictions. The runtime axis is also where the $10$ FPS setting matters most: a $10$ Hz control loop must run at the data rate and \emph{both} frontier models clear it, $\pi_0$ at $15.9$ Hz and SmolVLA at $31.6$ Hz. SmolVLA's effective rate is boosted by its $4$-step action chunking, which runs the network once per four emitted actions, yielding a bimodal latency (median $3.2$ ms, p90 $122$ ms); $\pi_0$ is steady ($62.8\pm1.4$ ms). A practitioner can therefore pick an operating point $1$ FPS for compute-frugal high-level guidance, $10$ FPS for reactive following with a frontier model that meets the corresponding control budget at each.

\begin{table}[t]
\renewcommand{\arraystretch}{1.2}
\centering
\caption{Single-frame inference latency on one H100 ($10$ warm-up, $50$ timed). ``Rate'' is sustained throughput; the last block is the $10$ FPS frontier.}
\label{tab:latency}
\setlength{\tabcolsep}{4pt}
\small
\begin{tabular}{llrrrc}
\toprule
\textbf{Rate} & \textbf{Model} & \textbf{Params} & \textbf{Lat.\ (ms)} & \textbf{p95} & \textbf{Hz} \\
\midrule
\multirow{4}{*}{1 FPS}
 & SmolVLA       & $0.45$B & $\mathbf{36.5}$ & $124$ & $\mathbf{27.4}$ \\
 & $\pi_0$/OpenPI& $3.3$B  & $67.2$ & $70$  & $14.9$ \\
 & GR00T~N1.7    & $3.1$B  & $77.9$ & $101$ & $12.8$ \\
 & OpenVLA-7B    & $7.6$B  & $365.1$& $376$ & $2.7$ \\
\midrule
\multirow{2}{*}{10 FPS}
 & SmolVLA       & $0.45$B & $31.7^{\ast}$ & $122$ & $\mathbf{31.6}$ \\
 & $\pi_0$/OpenPI& $3.3$B  & $62.8$ & $65$  & $15.9$ \\
\bottomrule
\multicolumn{6}{l}{\footnotesize $^{\ast}$bimodal (median $3.2$ ms / p90 $122$ ms) from $4$-step chunking.}
\end{tabular}
\end{table}

\begin{figure}[tbp]
\centering
\includegraphics[width=\linewidth,height=0.72\textheight,keepaspectratio]{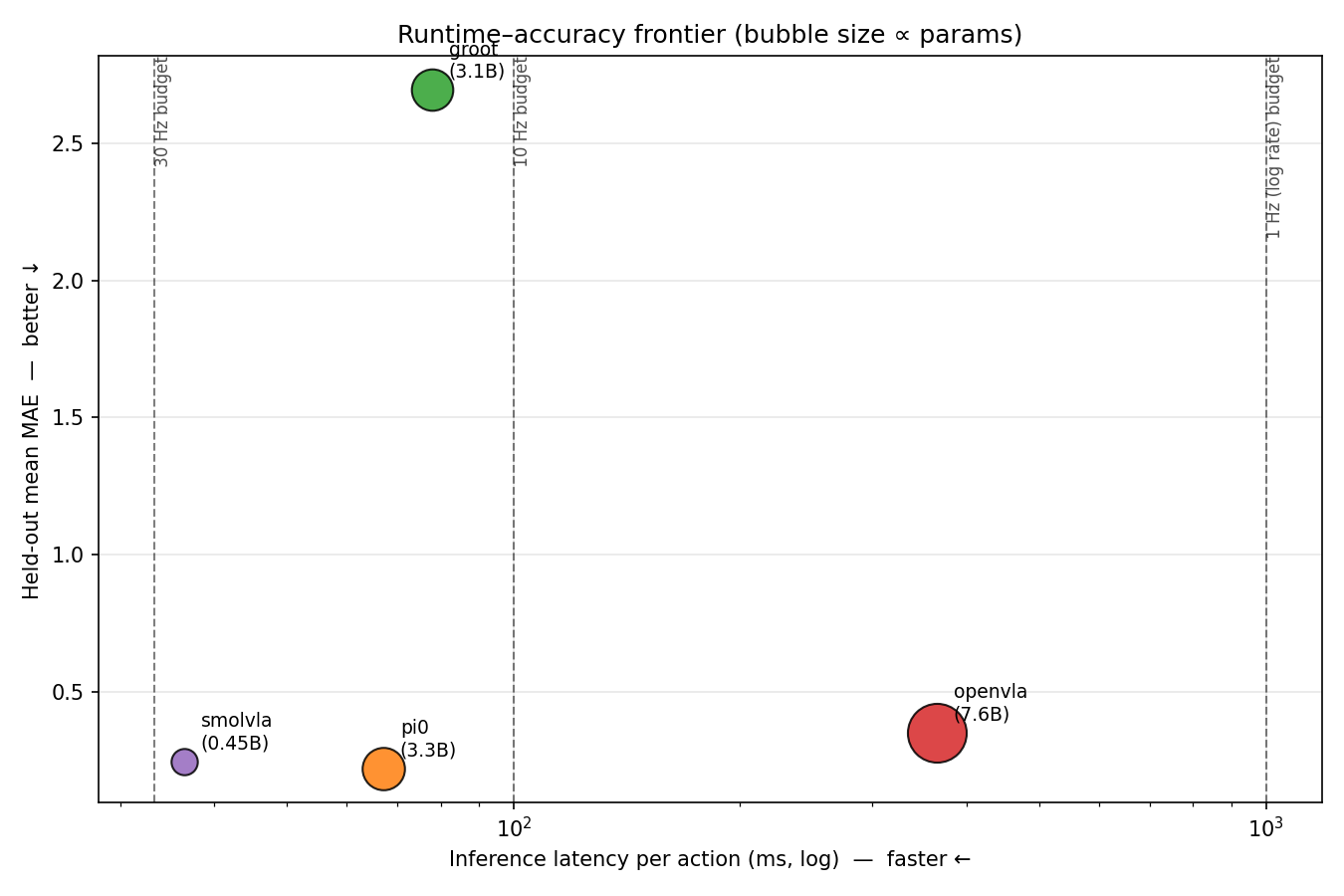}
\caption{Runtime--accuracy frontier. $x$-axis: single-frame inference latency (log ms, faster$\leftarrow$); $y$-axis: held-out mean MAE (lower is better); bubble area $\propto$ parameter count. Vertical dashed lines mark the per-action time budgets for $1$, $10$ and $30$ Hz control loops. SmolVLA and $\pi_0$ sit in the accurate-and-fast quadrant and clear the $10$ Hz budget; OpenVLA-7B violates even the $10$ Hz budget and GR00T is fast but inaccurate.}
\label{fig:latency_accuracy}
\end{figure}

\subsection{Open-loop rollout drift}
\label{sec:rollout}

To complement the per-step view with a compounding one, we integrate each predicted action sequence forward from the true initial pose of every held-out episode and measure Average and Final Displacement Error (ADE/FDE) against the expert path (Table \ref{tab:rollout}). The compounding view sharpens and \emph{confirms} the ranking at \emph{both} rates: $\pi_0$ drifts least ($1$ FPS $2.55/4.73$ m; $10$ FPS $3.12/5.64$ m), SmolVLA follows ($3.06/5.95$; $3.74/7.00$) and at $1$ FPS OpenVLA and especially GR00T trail far behind. FDE exceeds ADE for every model, reflecting accumulating drift on curved segments where the unpredicted lateral and heading corrections compound. Critically, rollout drift separates $\pi_0$ from SmolVLA \emph{cleanly at $10$ FPS} precisely where per-step MAE saturates against the baseline which is why we recommend reporting closed-loop rollout drift and correlation, not raw per-step MAE alone, when evaluating at a deployment frame rate.

\begin{table}
\renewcommand{\arraystretch}{1.2}
\centering
\caption{Open-loop rollout error (m) integrated from the true initial pose. ADE/FDE: average/final displacement error. The model ranking holds at both rates.}
\label{tab:rollout}
\setlength{\tabcolsep}{5pt}
\small
\begin{tabular}{llrr}
\toprule
\textbf{Rate} & \textbf{Model} & \textbf{ADE} & \textbf{FDE} \\
\midrule
\multirow{4}{*}{1 FPS}
 & $\pi_0$/OpenPI & $\mathbf{2.55}$ & $\mathbf{4.73}$ \\
 & SmolVLA        & $3.06$ & $5.95$ \\
 & OpenVLA-7B     & $3.56$ & $6.30$ \\
 & GR00T~N1.7     & $24.92$ & $43.84$ \\
\midrule
\multirow{2}{*}{10 FPS}
 & $\pi_0$/OpenPI & $\mathbf{3.12}$ & $\mathbf{5.64}$ \\
 & SmolVLA        & $3.74$ & $7.00$ \\
\bottomrule
\end{tabular}
\end{table}

\begin{figure}[tbp]
\centering
\includegraphics[width=\linewidth,height=0.72\textheight,keepaspectratio]{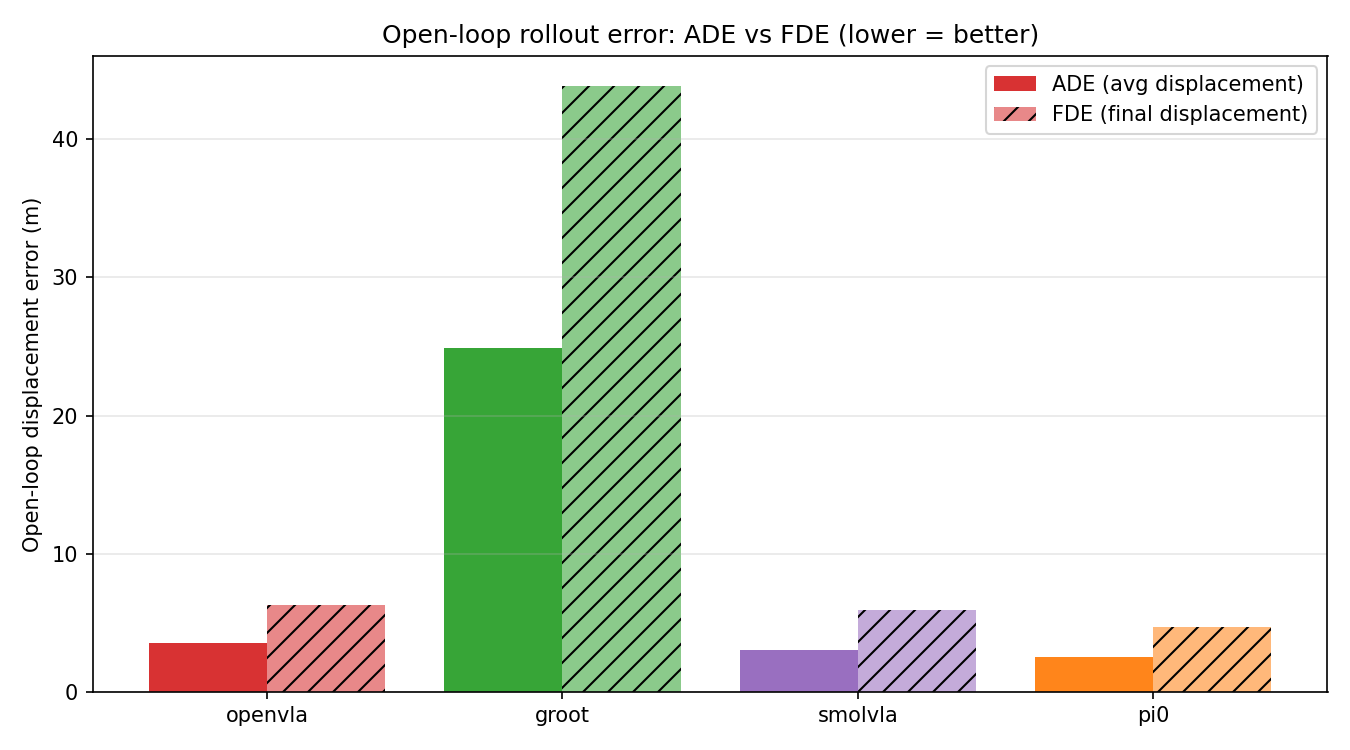}
\caption{Open-loop rollout error at $1$ FPS: average (ADE, solid) vs.\ final (FDE, hatched) displacement error from the true initial pose. The flow-matching models drift least; OpenVLA trails; GR00T's normalization failure produces order-of-magnitude larger drift. FDE exceeds ADE for every model, the signature of compounding error.}
\label{fig:ade_fde}
\end{figure}

\begin{figure}[tbp]
\centering
\includegraphics[width=\linewidth,height=0.72\textheight,keepaspectratio]{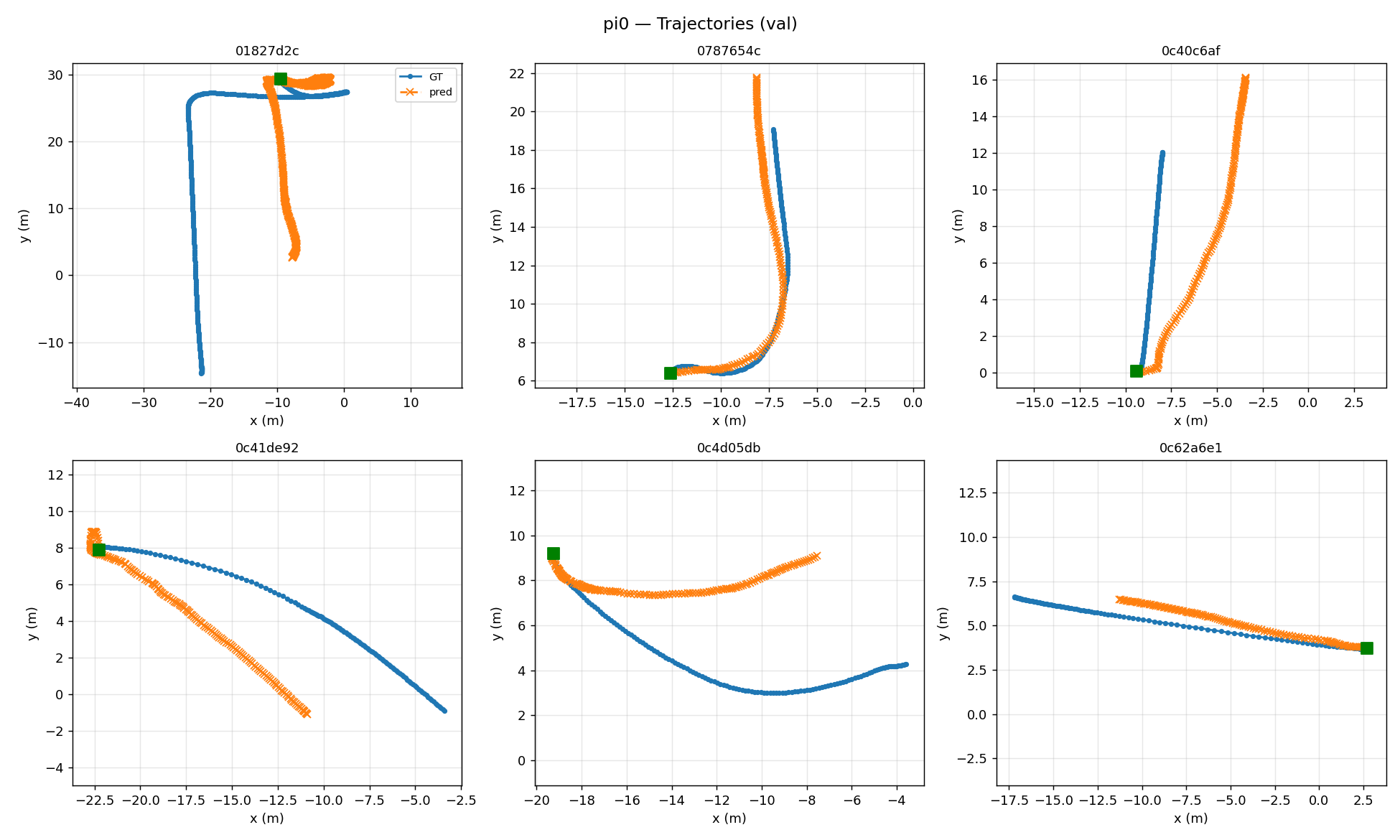}
\caption{$\pi_0$ open-loop rollouts (orange) vs.\ expert paths (blue) on six $10$ FPS held-out episodes, integrated from the true initial pose (green square). Rollouts capture the gross forward progression but accumulate drift on curved segments where the unpredicted lateral and heading corrections compound the trajectory-level manifestation of the off-axis channels' low correlation.}
\label{fig:traj_10fps}
\end{figure}

\section{Discussion}
\label{sec:discussion}

\begin{widefigure}[tbp]
\centering
\definecolor{cCoreLine}{RGB}{52,73,94}   \definecolor{cCoreFill}{RGB}{237,241,245}
\definecolor{cJepaLine}{RGB}{31,97,160}  \definecolor{cJepaFill}{RGB}{216,233,248}
\definecolor{cDynLine}{RGB}{16,124,106}  \definecolor{cDynFill}{RGB}{209,239,230}
\definecolor{cReasLine}{RGB}{191,119,10} \definecolor{cReasFill}{RGB}{253,238,213}
\definecolor{cPlanLine}{RGB}{106,61,154} \definecolor{cPlanFill}{RGB}{233,224,246}
\definecolor{cPhysLine}{RGB}{176,48,50}  \definecolor{cPhysFill}{RGB}{250,223,221}
\definecolor{cEvalLine}{RGB}{77,124,47}  \definecolor{cEvalFill}{RGB}{229,241,215}
\resizebox{0.99\textwidth}{!}{
\begin{tikzpicture}[font=\footnotesize, >=Latex, line cap=round, line join=round]
\tikzset{
  module/.style 2 args={rounded corners=2.2mm, draw=#1, fill=#2, line width=1.1pt,
        align=left, text width=4.05cm, inner sep=7pt},
  badge/.style={circle, draw=white, line width=0.8pt, fill=#1, text=white,
        font=\bfseries\small, inner sep=0.5pt, minimum size=5.2mm},
  flow/.style={-{Latex[length=2.6mm,width=2.0mm]}, line width=1.15pt, draw=#1},
  fb/.style={-{Latex[length=2.4mm,width=1.9mm]}, line width=1.05pt, draw=#1,
        dash pattern=on 3.2pt off 2.4pt},
  elab/.style={font=\scriptsize, text=#1, fill=white, rounded corners=1pt,
        inner sep=1.6pt, align=center},
  evalflow/.style={-{Latex[length=2.2mm,width=1.7mm]}, line width=0.9pt, draw=black!45}
}

\foreach \sx in {-1,1}{
  \draw[line width=2.6pt, cCoreLine!85] (\sx*1.9, 0.9) -- (\sx*4.0, 3.3);
  \draw[line width=2.6pt, cCoreLine!85] (\sx*1.9, -0.1) -- (\sx*4.0, -2.5);
}

\foreach \x/\y/\s in {-4/3.3/1, 4/3.3/-1, -4/-2.5/-1, 4/-2.5/1}{
  \draw[cCoreLine!45, line width=0.7pt, dash pattern=on 2.4pt off 2.2pt] (\x,\y) circle (0.82);
  \draw[fill=cCoreLine!25, draw=cCoreLine!75, line width=0.5pt,
        rotate around={40:(\x,\y)}] (\x,\y) ellipse (0.70 and 0.13);
  \draw[fill=cCoreLine!25, draw=cCoreLine!75, line width=0.5pt,
        rotate around={-40:(\x,\y)}] (\x,\y) ellipse (0.70 and 0.13);
  \fill[cCoreLine] (\x,\y) circle (0.14);
  \draw[-{Latex[length=1.5mm,width=1.2mm]}, cCoreLine!60, line width=0.6pt]
        ($(\x,\y)+({90-\s*38}:1.04)$) arc[start angle={90-\s*38}, end angle={90+\s*38}, radius=1.04];
}
\node[rounded corners=3.2mm, draw=cCoreLine, fill=cCoreFill, line width=1.5pt,
      align=center, text width=4.6cm, inner sep=7pt] (core) at (0,0.4)
  {\textcolor{cCoreLine}{\textbf{WareFly-VLA Data Core}}\\[2pt]
   {\scriptsize RGB video $\cdot$ appearance-grounded\\
    instructions $\cdot$ continuous 4-DoF actions\\
    ground-truth pose $\cdot$ difficulty tags}};

\node[module={cJepaLine}{cJepaFill}] (jepa) at (-10.3,4.6) {
  \textcolor{cJepaLine}{\textbf{JEPA predictive representation}}\\[2pt]
  {\scriptsize \textbullet~Predict future \emph{embeddings}, not pixels\\
  \textbullet~Compact state $z_t$: target identity $+$ relative geometry\\
  \textbullet~Identity persistence under occlusion (\emph{obstructed}, \emph{find} tags)\\[1pt]
  \cite{lecun2022path,assran2023ijepa,bardes2024vjepa}}};

\node[module={cDynLine}{cDynFill}] (dyn) at (0,6.9) {
  \textcolor{cDynLine}{\textbf{Action-conditioned world dynamics}}\\[2pt]
  {\scriptsize \textbullet~Latent dynamics $p(z_{t+1}\mid z_t,\mathbf{a}_t,\ell)$\\
  \textbullet~Pose-verified rollouts with uncertainty\\
  \textbullet~Recovers relative velocity: the $\dyl$/$\dz$ bottleneck\\[1pt]
  \cite{ha2018worldmodels,hafner2019planet,hansen2024tdmpc2,bruce2024genie}}};

\node[module={cReasLine}{cReasFill}] (reas) at (10.3,4.6) {
  \textcolor{cReasLine}{\textbf{Language-conditioned prediction \& reasoning}}\\[2pt]
  {\scriptsize \textbullet~Instructions as constraints on future states\\
  \textbullet~Phase prediction: search $\to$ acquire $\to$ follow\\
  \textbullet~Counterfactual target hypotheses\\[1pt]
  \cite{tian2025uavllm,cogdrone2025,deshpande2026airnav}}};

\node[module={cPlanLine}{cPlanFill}] (plan) at (10.3,-4.2) {
  \textcolor{cPlanLine}{\textbf{Model-based planning \& policy synthesis}}\\[2pt]
  {\scriptsize \textbullet~Latent MPC: score actions before flight\\
  \textbullet~Imagination-based reinforcement learning\\
  \textbullet~Distill plans into flow-matching VLA heads\\[1pt]
  \cite{hafner2020dreamer,hafner2023dreamerv3,assran2025vjepa2,black2024pi0}}};

\node[module={cPhysLine}{cPhysFill}] (phys) at (-10.3,-4.2) {
  \textcolor{cPhysLine}{\textbf{Physical AI \& sim-to-real transfer}}\\[2pt]
  {\scriptsize \textbullet~Physics-calibrated simulation prior\\
  \textbullet~Domain randomization $+$ generative scene augmentation\\
  \textbullet~Safety envelopes for human-proximal flight\\[1pt]
  \cite{tobin2017domainrandomization,nvidia2025cosmos,kaufmann2023champion,pegasus_sim}}};

\node[badge=cJepaLine] at (jepa.north west) {1};
\node[badge=cDynLine]  at (dyn.north west)  {2};
\node[badge=cReasLine] at (reas.north east) {3};
\node[badge=cPlanLine] at (plan.north east) {4};
\node[badge=cPhysLine] at (phys.north west) {5};

\draw[flow=cJepaLine] ($(-4,3.3)+(160:0.84)$) to[out=160,in=-25]
  node[elab=cJepaLine, pos=0.5] {$E1$: video context} (jepa.east);
\draw[flow=cDynLine] (core.north) --
  node[elab=cDynLine, pos=0.5] {$E2$: state--action--pose tuples} (dyn.south);
\draw[flow=cReasLine] ($(4,3.3)+(20:0.84)$) to[out=20,in=205]
  node[elab=cReasLine, pos=0.5] {$E3$: instructions $+$ tags} (reas.west);

\draw[flow=cJepaLine] (jepa.north) to[out=35,in=180]
  node[elab=cJepaLine, pos=0.5] {$E4$: latent state $z_t$} (dyn.west);
\draw[flow=cDynLine, rounded corners=5mm] (dyn.east) -- (13.55,6.9) -- (13.55,-4.2) -- (plan.east);
\node[elab=cDynLine, rotate=90] at (13.55,1.35) {$E5$: imagined rollouts $\hat z_{t+1:t+H}$};
\draw[flow=cReasLine] (reas.south) --
  node[elab=cReasLine, pos=0.5] {$E6$: goals $+$ constraints} (plan.north);

\draw[flow=cPlanLine] (plan.west) --
  node[elab=cPlanLine, pos=0.5] {$E7$: candidate action plans} (phys.east);
\draw[fb=cPhysLine] (phys.north) to[out=70,in=190]
  node[elab=cPhysLine, pos=0.5] {$E8$: deployment gaps $\to$ new data} (core.west);

\node[rounded corners=2mm, draw=cEvalLine, fill=cEvalFill, line width=1.0pt,
      align=center, text width=12.8cm, inner sep=5.5pt] (eval) at (0,-7.35) {
  \textcolor{cEvalLine}{\textbf{Evaluation hooks in the released benchmark}}\\[1.5pt]
  {\scriptsize occlusion-recovery rate $\cdot$ rollout drift (ADE/FDE) $\cdot$ same-rate skill
   score $s_d$ $\cdot$ instruction-paraphrase suites $\cdot$ safety-margin compliance}};
\draw[evalflow, rounded corners=2.5mm] (plan.south) |- (eval.east);
\draw[evalflow, rounded corners=2.5mm] (phys.south) |- (eval.west);
\end{tikzpicture}}
\caption{World-model research roadmap enabled by WareFly-VLA. The dataset core is drawn as a quadrotor and aligns five synchronized supervision streams. Three predictive modules consume these streams: JEPA-style representation learning (1), action-conditioned latent dynamics (2) and language-conditioned prediction and reasoning (3). Their outputs fuse into model-based planning (4), which reaches hardware through Physical-AI and sim-to-real transfer (5). Solid colored arrows ($E1$--$E7$) denote forward modeling and control dependencies; the dashed arrow ($E8$) returns deployment failures to data collection and closes the loop. The bottom band lists evaluation hooks already provided by the benchmark release.}
\label{fig:world_model_taxonomy}
\end{widefigure}

The benchmark reveals three consistent characteristics of current Vision--Language--Action policies. First, flow-matching policies ($\pi_0$ and SmolVLA) consistently outperform the discrete-token formulation adopted by OpenVLA. This advantage is primarily attributed to the continuous nature of flow matching, which avoids the quantization error introduced by discretizing low-amplitude UAV actions. The effect is particularly evident for the altitude and heading channels, where the empirical action range is considerably smaller than the discretization resolution. These observations are consistent with recent studies advocating continuous action representations for embodied control \cite{kim2025oft,pertsch2025fast}; inference-time steering of pre-trained VLAs \cite{li2026vlapilot} is a complementary route to narrowing the held-out gap on small aerial corpora without additional fine-tuning.

Second, the benchmark exposes an inherent limitation of single-frame aerial perception. While forward motion ($\Delta x$) can be estimated from instantaneous visual cues such as target scale and relative position, lateral ($\Delta y$) and vertical ($\Delta z$) corrections depend largely on temporal information describing target motion. Consequently, all evaluated policies exhibit near-zero correlation on these channels, indicating that the bottleneck arises primarily from limited temporal observability rather than insufficient model capacity.

Finally, evaluation across different sampling rates demonstrates that per-step MAE alone is sensitive to frame rate. Increasing the sampling frequency reduces the magnitude of ground-truth actions, allowing even a constant-mean predictor to achieve deceptively low absolute error. Correlation, in contrast, remains comparatively stable because it measures agreement in temporal dynamics rather than action magnitude. These observations suggest that future aerial VLA benchmarks should complement per-step MAE with correlation-based measures, normalized skill scores, and closed-loop trajectory evaluation to provide a more comprehensive assessment of policy performance.

\subsection{World-model roadmap enabled by WareFly-VLA}
\label{sec:world_model_roadmap}

Our results expose a structural limit of reactive VLA policies. A single frame does not reveal relative motion, so the lateral and vertical channels remain at chance. World models target this limit directly. They maintain an internal state that summarizes past observations, predict how this state evolves under candidate actions and thereby support planning before acting \cite{ha2018worldmodels,lecun2022path}. WareFly-VLA aligns five synchronized supervision streams for this purpose: RGB video, appearance-grounded instructions, continuous $4$-DoF actions, ground-truth pose and difficulty tags. Fig. \ref{fig:world_model_taxonomy} organizes the resulting research space as a directed graph around the dataset core, with eight dependency edges ($E1$--$E8$) grouped into supervision ($E1$--$E3$), prediction--planning fusion ($E4$--$E6$) and a closed deployment loop ($E7$--$E8$). We highlight five directions.

\emph{(1) JEPA-style predictive representation ($E1$, $E4$).} Joint-embedding predictive architectures learn a latent state $z_t$ by predicting embeddings of future observations rather than pixels \cite{lecun2022path,assran2023ijepa,bardes2023mcjepa,bardes2024vjepa}. This objective discards warehouse texture and retains task-relevant factors such as target identity and relative geometry. The \emph{obstructed} and \emph{find} tags provide direct probes of identity persistence under occlusion, the failure case that most affects person-following.

\emph{(2) Action-conditioned latent dynamics ($E2$, $E5$).} The synchronized action--pose stream supports learning a latent flight-dynamics model $p(z_{t+1}\mid z_t,\mathbf{a}_t,\ell)$ conditioned on the instruction $\ell$ \cite{hafner2019planet,hansen2024tdmpc2}. Ground-truth pose gives a physical consistency check: imagined rollouts must agree with true ego-motion. Such a model recovers target-relative velocity from temporal context, which is exactly the information the single-frame bottleneck removes. Action-conditioned video generation offers a complementary, pixel-space route to the same capability \cite{bruce2024genie,hu2023gaia}.

\emph{(3) Language-conditioned prediction and reasoning ($E3$, $E6$).} Instructions in WareFly-VLA constrain \emph{who} to follow, not \emph{where} to go. A predictive model can treat them as constraints on future states: which worker will match the description and where that worker will be. Concrete subtasks include phase prediction over the search--acquire--follow structure of Fig. \ref{fig:teleop_policy} and counterfactual target hypotheses during active search \cite{tian2025uavllm,cogdrone2025}.

\emph{(4) Model-based planning and policy synthesis ($E5$--$E7$).} Given a latent dynamics model, the UAV can score candidate action sequences before executing them, through latent model-predictive control or imagination-based learning \cite{hafner2020dreamer,hafner2023dreamerv3}. Recent self-supervised video world models already support such zero-shot robot planning \cite{assran2025vjepa2}. Planned behaviors can then be distilled into fast flow-matching heads such as $\pi_0$ and SmolVLA \cite{black2024pi0,shukor2025smolvla}, so deployment stays within the latency budgets of Section \ref{sec:latency}.

\emph{(5) Physical AI and sim-to-real transfer ($E7$, $E8$).} The corpus is generated under the physics stack of Isaac Sim and Pegasus \cite{NVIDIA_Isaac_Sim,pegasus_sim}, so learned dynamics inherit a calibrated simulator prior. Domain randomization and world-foundation-model scene augmentation can widen the visual and physical distribution \cite{tobin2017domainrandomization,nvidia2025cosmos}, while agile-flight results show that simulation-trained aerial policies can reach real hardware \cite{kaufmann2023champion}. Deployment failures return to the data core as new episodes ($E8$), closing the loop.

Each direction is testable with hooks already in the release: occlusion-recovery rate on tagged episodes, rollout drift (ADE/FDE, Table \ref{tab:rollout}). The roadmap thus turns WareFly-VLA from a benchmark of reactive policies into a substrate for predictive, physically grounded aerial embodied intelligence.

\subsection{Limitations}
\label{sec:limitations}
Four limitations bound the scope of our conclusions. \emph{Open-loop evaluation:} every policy is scored per step against the ground-truth frame rather than the frame its own past actions would induce; closed-loop deployment in simulation or on hardware an essential follow-up is expected to \emph{amplify} the reported gaps, since compounding drift penalizes weaker predictors super-linearly. \emph{Dataset scale:} the corpus is $507$ teleoperated episodes ($8{,}504$ non-terminal transitions), small by manipulation-VLA standards; this deliberately trades raw count for task fidelity and linguistic diversity, so absolute saturation will require larger collections. \emph{Single environment and simulation:} all episodes come from one photorealistic Isaac~Sim warehouse, so cross-environment and sim-to-real transfer are not assessed here; a second-warehouse extension is in progress. \emph{Single-frame, single-camera observation:} our protocol uses one forward-facing RGB stream, which is precisely what bottlenecks the off-axis channels analysed above; multi-frame, multi-view or RGB-D inputs are a natural extension and would also admit backbones designed for richer observations (\emph{e.g.}, RDT-1B's two-camera setup \cite{liu2024rdt}).

\section{Conclusion}
\label{sec:conclusion}

We have presented \textbf{WareFly-VLA}, a Vision--Language--Action dataset and benchmark for language-conditioned drone person-following in a photorealistic industrial warehouse. The benchmark evaluates four open-source VLA architectures spanning two orders of magnitude in parameter count. All models share one canonical, leakage-free episode-level split, bootstrap confidence intervals and two control rates ($1$ and $10$ FPS).

The study yields five findings of broad relevance to the VLA and aerial-robotics communities. First, the task is far from solved. The strongest model, $\pi_0$, reaches only $r{\approx}0.5$ on the forward and heading channels and no model beats a constant-mean baseline on the lateral or vertical channels, which lack single-frame signal. Second, every model shows a large train-to-validation gap and the smallest model overfits most. Held-out capacity, not training fit or size alone, determines the ranking. Third, continuous flow-matching action heads hold a consistent, statistically supported edge over discrete action tokenization, which collapses low-amplitude altitude and yaw commands. Fourth, the GR00T~N1.7 new-embodiment interface fails to recalibrate its normalization statistics under limited-data aerial transfer. This failure mode should inform future foundation-model design. Fifth, the architecture ranking holds at both control rates, but per-step MAE loses discriminative power as the frame rate rises. Correlation and closed-loop rollout drift are therefore the rate-robust metrics for deployment-oriented aerial VLA evaluation.

Beyond the benchmark, we chart a world-model research roadmap, from JEPA-style predictive representation to Physical-AI transfer, that the dataset's synchronized supervision streams enable. We release the dataset, baseline implementations and analysis code under an open license. We invite the community to extend the benchmark with additional baselines (\emph{e.g.}, OpenVLA-OFT \cite{kim2025oft}, $\pi_0$-FAST \cite{pertsch2025fast}, RDT-1B \cite{liu2024rdt} and Octo \cite{octo2024}), additional warehouse environments and closed-loop hardware deployment. We hope WareFly-VLA catalyzes progress toward safe and capable language-guided aerial autonomy in real industrial environments.

\section*{CRediT authorship contribution statement}

\section*{Declaration of competing interest}
The authors declare that they have no known competing financial interests or personal relationships that could have appeared to influence the work reported in this paper.

\section*{Data availability}
The WareFly-VLA dataset, benchmark splits, training pipelines and evaluation scripts are available at \url{https://huggingface.co/datasets/SIMOGroup/WareFly-VLA} (access on request during review; public release upon acceptance).

\section*{Declaration of generative AI and AI-assisted technologies in the writing process}
During the preparation of this work the authors used Grammarly (professional version) in order to improve the readability and language of the manuscript. After using this tool, the authors reviewed and edited the content as needed and take full responsibility for the content of the published article.

\section*{Acknowledgments}
This work was supported by the project entitled ``BioDroneX: AI-Enhanced Bio-Inspired Drone for Adaptive Multi-Environment Missions'' at the Center for AI Research, VinUniversity.

\appendix
\setcounter{figure}{0}
\setcounter{table}{0}
\section*{Appendices}
The following appendices collect the reproducibility protocol, the worker-appearance and PPE inventory, the failure-mode analysis, additional robustness and cross-model analyses, and extended $10$~FPS results referenced from the main text.

\section{Reproducibility Protocol}
\label{app:repro}
The canonical leakage-free split is built once and serialised to a manifest so that every per-model converter consumes the identical training and validation episode sets. The recipe is:
\begin{enumerate}
\item Enumerate all $507$ episode UUIDs and sort them lexicographically (so the partition is invariant to insertion order and to filesystem listing order).
\item Seed a NumPy generator with the integer $0$ and apply a permutation.
\item Assign the first $76$ episodes ($15\%$) to the validation set and the remaining $431$ to the training set.
\item Write a JSON file listing the two episode ID sets; every model adapter (LeRobot, OpenVLA, GR00T, OpenPI/$\pi_0$) reads this file when converting the raw corpus into its own format.
\item Exclude terminal frames (those whose action labels are identically zero by construction) from both training and evaluation.
\end{enumerate}

Under a naive \emph{train-on-everything} protocol the flow-matching models report near-perfect held-out correlations ($r{>}0.94$ on every dimension), a number that collapses to the values reported in Table \ref{tab:main_results} once the leakage above is removed. We emphasise this because the gap between the two protocols \emph{reverses} the model ranking and any future submission to the WareFly-VLA benchmark should adopt the leakage-free split verbatim.

\begin{table}[t]
\renewcommand{\arraystretch}{1.2}
\centering
\caption{Per-model fine-tuning hyperparameters used in the benchmark. All runs use a single H100~$80$GB GPU. Total parameters / trainable parameters appear in Table \ref{tab:baselines}.}
\label{tab:hparams}
\setlength{\tabcolsep}{4pt}
\begin{tabular}{lcccc}
\toprule
 & \textbf{OpenVLA} & \textbf{GR00T} & \textbf{$\pi_0$} & \textbf{SmolVLA} \\
\midrule
Adaptation     & LoRA & proj.+DiT & LoRA & full \\
LoRA $r/\alpha$& $16/32$ &   & expert head &   \\
Optimiser      & AdamW & AdamW & AdamW & AdamW \\
Learning rate  & $2{\times}10^{-4}$ & $1{\times}10^{-4}$ & $1{\times}10^{-4}$ & $5{\times}10^{-5}$ \\
Steps / epochs & $20$ ep & $6{,}000$ st & $15{,}000$ st & $15{,}000$ st \\
Batch size     & $4$ & $16$ & $16$ & $32$ \\
Precision      & 4-bit & bf16 & bf16 & bf16 \\
Input image    & $224^2$ & $256^2$ & $256^2$ & $256^2$ \\
\bottomrule
\end{tabular}
\end{table}

\emph{Latency.} Single-frame inference latency is reported in the runtime experiments through an opt-in benchmark mode that times only the observation-to-action call after $10$ warm-up iterations, over $50$ timed iterations, with explicit CUDA synchronisation between calls; the reported median and p95 therefore measure model forward cost rather than data-loading or simulator-update overhead.

\section{Worker Appearance and PPE Inventory}
\label{app:appearance}
The simulated workers used as targets are dressed from a curated personal-protective-equipment (PPE) inventory designed to maximise lexical grounding diversity while remaining realistic for an industrial-warehouse setting. The inventory is grouped into six categories:
\begin{itemize}
\item \emph{Headwear:} hard hats (orange, yellow, white, blue, red), bump caps, beanies and uncovered hair (varied colours including the corpus's \emph{red hair} mentions).
\item \emph{Upper body:} high-visibility safety vests (orange, yellow, lime, with and without reflective tape), button-up shirts (long- and short-sleeve, in colours such as terracotta, teal, moss green, lavender, purple), polo shirts and t-shirts.
\item \emph{Lower body:} cargo trousers, work pants, overalls and coveralls in neutral and high-visibility colours.
\item \emph{Footwear:} steel-toe work boots and industrial trainers (typically dark brown or black).
\item \emph{Hand protection:} cut-resistant gloves, leather gloves and ungloved variants.
\item \emph{Accessories:} ear-protection headphones, dust masks, safety glasses, lanyards and clip-on identification tags.
\end{itemize}
Each episode samples one item from each applicable category, subject to the constraint that no two episodes share an identical configuration. This combinatorial structure is what supports the rich appearance vocabulary documented in the linguistic analysis (Section \ref{sec:analysis}) and it is also what makes the \emph{find} difficulty regime non-trivial: distractor workers in the same scene typically share two or three attribute categories with the target while differing in the remaining ones.

\section{Failure Mode Analysis}
\label{app:failure}
We document three recurring failure modes that emerge from the per-sample scatter plots and the qualitative arrow overlays:
\begin{enumerate}
\item \emph{Forward-magnitude compression.} On the $\dxl$ channel, every working model produces predictions whose magnitudes are compressed toward the training-mean $\bar a_{\dxl}{=}0.864$ m/frame. Long-range pursuits at the right tail of the empirical distribution (up to $5.62$ m/frame, Table \ref{tab:action_stats}) are under-predicted with a clear range-compression slope ($\hat a {\approx}0.5 a$ in the large-action regime). This is the dominant residual error on the only learnable channel.
\item \emph{Off-axis collapse.} On the $\dyl$ and $\dz$ channels, predictions collapse toward the training mean band: $\dyl$ predictions cluster near zero across the full empirical range and $\dz$ predictions are effectively constant. This produces a low MAE that matches the constant-mean baseline a benign-looking aggregate metric that masks the fact that the model has not learned the channel at all.
\item \emph{Verb-token shortcut absence.} Per-verb-family error (Table \ref{tab:verb_breakdown}) shows no spike on rare verbs or on the heterogeneous \emph{other} bucket, confirming that the models do not exploit the task-initiating verb as a shortcut. This is the absence of a failure mode that we deliberately designed against in the paraphrase-diverse annotation protocol (Section \ref{sec:dataset}).
\end{enumerate}

The first two failure modes admit the same prescription: the channels that the per-frame policy cannot recover are precisely those whose ground-truth supervision encodes \emph{temporal} target-relative state (target velocity, target altitude change). Multi-frame observation or explicit velocity-from-pose conditioning is the route to closing them; we release the per-frame ground-truth pose state alongside the corpus to enable exactly this line of work.

\section{Additional Robustness and Cross-Model Analyses}
\label{app:robustness}
This appendix collects the robustness and cross-model diagnostics referenced from Sections \ref{sec:results} and \ref{sec:deployment}. They corroborate the headline per-dimension results the model ranking $\pi_0 > \text{SmolVLA} > \text{OpenVLA} > \text{GR00T}$ and the forward-channel-only learnability at finer granularity, but are not required to follow the main argument.

\subsection{Training dynamics}

Fig. \ref{fig:training} shows the loss trajectories for OpenVLA, GR00T and $\pi_0$ over their respective fine-tuning runs. The SmolVLA does not emit a step-level log file. We observed the training loss to descend from $1.42$ at initialization to $0.083$ at convergence over a $\sim$$50$-minute run. Each curve corresponds to a different optimization objective cross-entropy over discrete action tokens for OpenVLA, denoising score-matching for GR00T and flow-matching velocity regression for $\pi_0$ and the absolute scales are therefore not directly comparable across models. What is comparable is the \emph{shape}: OpenVLA's curve exhibits a clear overfitting onset around epoch $3$ ($\sim$$300$ steps) that we attribute to the LoRA adapters saturating on the available data; that the surrogate diffusion loss does not surface; $\pi_0$ exhibits a long quiescent flow-matching plateau before a sharp drop near step $1000$ once the action expert aligns with the VLM hidden states a dynamic also reported in the original $\pi_0$ paper \cite{black2024pi0}.

\begin{widefigure}[tbp]
  \centering
  \includegraphics[width=\linewidth,height=0.72\textheight,keepaspectratio]{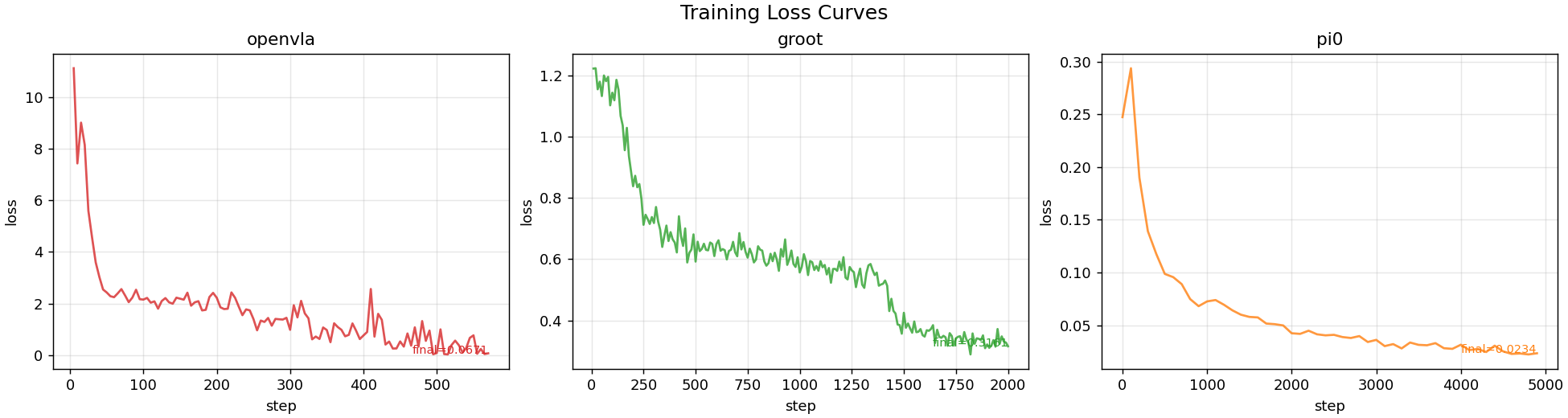}
  \caption{Training loss curves for the three models with step-level logs.
    OpenVLA optimizes cross-entropy over discrete action tokens; GR00T
    optimizes denoising score-matching; $\pi_0$ optimizes the flow-matching
    velocity objective.
    Annotations report the final loss value.}
  \label{fig:training}
\end{widefigure}

\subsection{All-model scatter comparison}

To complement the $\pi_0$ scatter in Fig. \ref{fig:scatter_pi0}, we show the equivalent held-out plots for the remaining three baselines (Figs. \ref{fig:scatter_smolvla}, \ref{fig:scatter_openvla}). SmolVLA reproduces $\pi_0$'s qualitative pattern a positive but range-compressed $\dxl$ trend with $\dyl$/$\dz$ collapsed to the prior band but with a steeper magnitude compression, consistent with its larger generalization gap. OpenVLA predictions exhibit a characteristic ``staircase'' artefact of discrete-token decoding: the $\dxl$ axis shows clear horizontal bands at quantization-bin boundaries and the $\dz$ axis collapses almost entirely to a single bin because the empirical $\dz$ range is smaller than the discretization grid resolution (\textit{cf.}\ Sec. \ref{sec:discussion}). GR00T predictions are dispersed two orders of magnitude outside the empirical action range, with no diagonal structure, consistent with the normalization failure diagnosed below.

\begin{widefigure}[tbp]
  \centering
  \includegraphics[width=\linewidth,height=0.72\textheight,keepaspectratio]{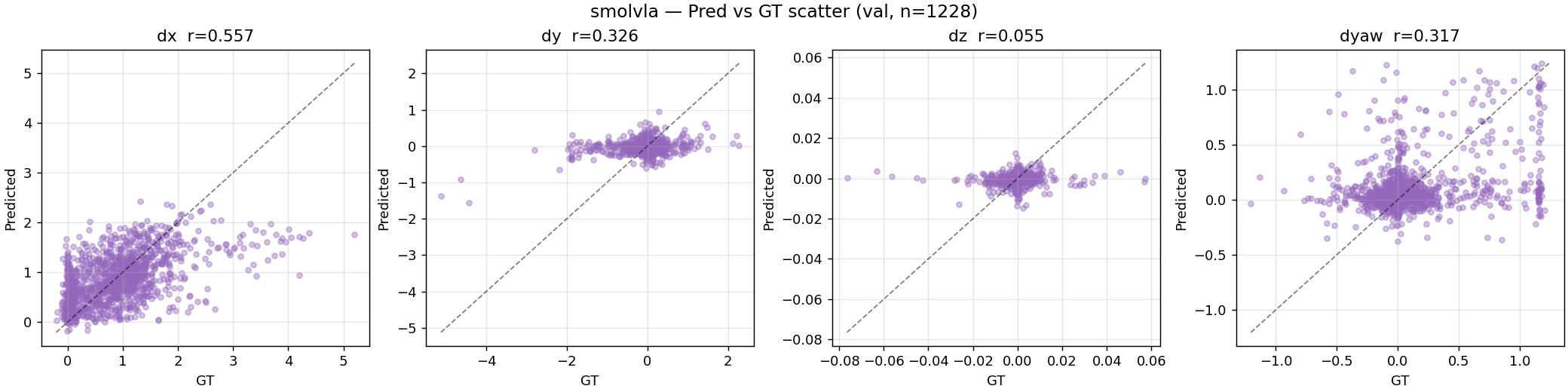}
  \caption{Per-dimension prediction-vs.-ground-truth scatter for
    \textbf{SmolVLA} on the held-out validation split.
    The dashed diagonal is the identity line; $\dxl$ is positive but
    range-compressed while $\dyl$ and $\dz$ collapse toward the prior mean.}
  \label{fig:scatter_smolvla}
\end{widefigure}

\begin{widefigure}[tbp]
  \centering
  \includegraphics[width=\linewidth,height=0.72\textheight,keepaspectratio]{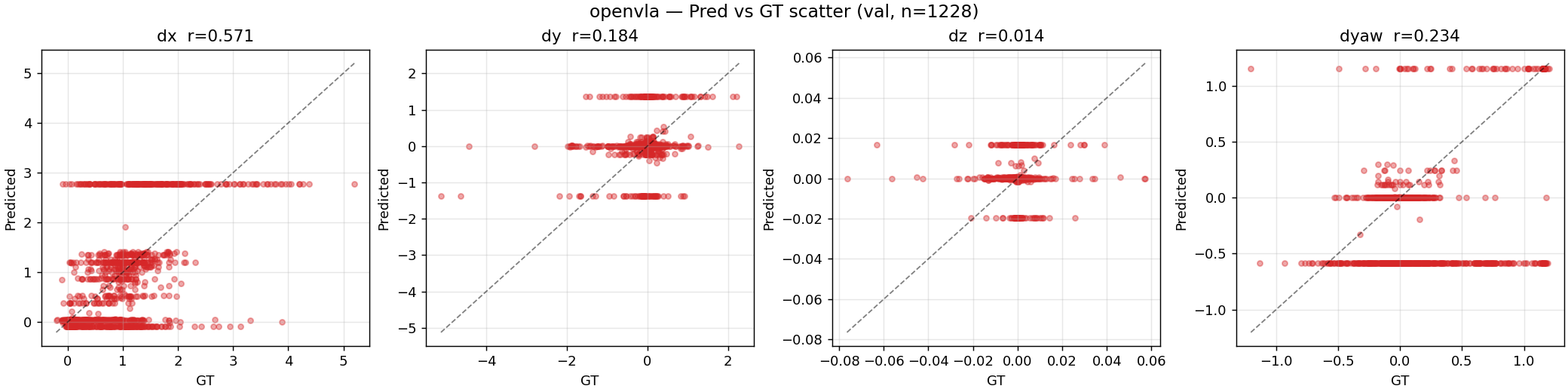}
  \caption{Per-dimension prediction-vs.-ground-truth scatter for
    \textbf{OpenVLA-7B} on the held-out split.
    Note the horizontal banding caused by $256$-bin discretization and the
    near-vertical collapse on $\dz$ where the discretization grid resolution
    exceeds the empirical action range.}
  \label{fig:scatter_openvla}
\end{widefigure}

\subsection{Error cumulative-distribution analysis}

Aggregate MAE conceals the shape of the error distribution; two models with identical MAE may differ substantially in their tail behavior, which is the most policy-relevant statistic for safety-critical UAV deployment. Fig. \ref{fig:cdf} plots the per-dimension cumulative distribution function (CDF) of $|{\rm err}|$ for each model on a symmetric-log $x$-axis. On the held-out split $\pi_0$ and SmolVLA exhibit nearly identical CDF shapes $\pi_0$ holds a small but consistent edge in the right tail of $\dxl$ and $\dpsi$ while OpenVLA's $\dxl$ CDF is shifted right by roughly $1.7\times$ (median $|err|$ on $\dxl$ near $0.73$ m versus $0.42$ m for the flow-matching models). GR00T's CDF lies one-to-two orders of magnitude above the empirical scale and is consistent with random-magnitude predictions. The most striking feature is shared by the three working models: the $\dyl$ and $\dz$ CDFs essentially coincide with the naive-baseline CDF, visually confirming that no model extracts held-out signal on the lateral and vertical channels.

\begin{widefigure}[tbp]
  \centering
  \includegraphics[width=\linewidth,height=0.72\textheight,keepaspectratio]{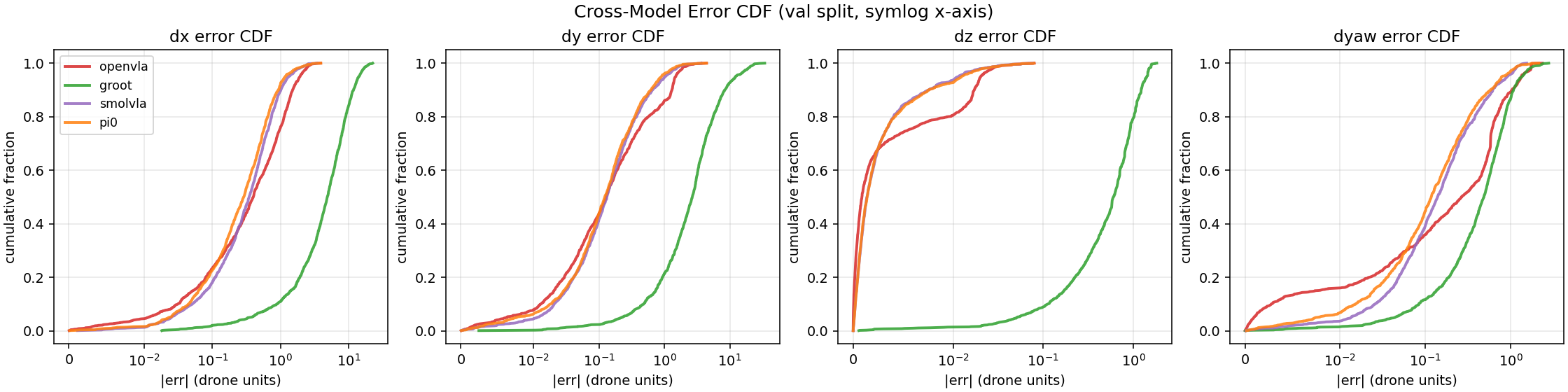}
    \caption{Cumulative distribution of per-dimension absolute error across all four models on the held-out validation split. The $x$-axis uses a symmetric log scale to preserve both small and large error magnitudes. $\pi_0$ (orange) and SmolVLA (purple) nearly coincide, with $\pi_0$ holding the right-tail edge; OpenVLA (red) is shifted right on $\dxl$; GR00T (green) lies far above the empirical scale. On $\dyl$/$\dz$ all working models track the naive baseline.}
  \label{fig:cdf}
\end{widefigure}

\subsection{Train--val gap analysis}

Table \ref{tab:train_val_gap} and Fig. \ref{fig:train_val} expand the generalization-gap analysis of Section \ref{sec:gen-gap} to every dimension. The gap is substantial for all models and counter to the intuition that a smaller model regularizes better it is \emph{largest} for the smallest model: SmolVLA's $\dxl$ MAE degrades from $0.083$ (train) to $0.468$ (val), a $+0.385$ m gap, versus $+0.263$ for $\pi_0$ and $+0.149$ for OpenVLA. SmolVLA attains the lowest training error of any model on every dimension yet the steepest fall-off on held-out episodes, the signature of a high-capacity action expert memorizing $431$ training trajectories. OpenVLA's gap is smallest in absolute terms only because it never fits the training data well to begin with, a discretization-induced underfit rather than good generalization. This decomposition overfitting for the flow-matching models, underfitting for the discrete-token model is the central diagnostic that the leakage-free protocol makes visible and that train-on-everything evaluation conceals.

\begin{table}[t]
\renewcommand{\arraystretch}{1.2}
\centering
\caption{Train vs. validation MAE per dimension (the gap quantifies overfitting). Models with no train predictions available are omitted.}
\label{tab:train_val_gap}
\begin{tabular}{lcccc}
\toprule
\textbf{Model} & $\Delta x_\ell$ & $\Delta y_\ell$ & $\Delta z$ & $\Delta\psi$ \\
\midrule
openvla (train) & 0.472 & 0.297 & 0.004 & 0.358 \\
openvla (val)   & 0.620 & 0.368 & 0.005 & 0.409 \\
\quad gap   & +0.149 & +0.070 & +0.001 & +0.052 \\
\midrule
smolvla (train) & 0.083 & 0.050 & 0.002 & 0.063 \\
smolvla (val)   & 0.468 & 0.264 & 0.003 & 0.243 \\
\quad gap   & +0.385 & +0.214 & +0.002 & +0.180 \\
\midrule
pi0 (train) & 0.150 & 0.081 & 0.002 & 0.084 \\
pi0 (val)   & 0.413 & 0.244 & 0.003 & 0.220 \\
\quad gap   & +0.263 & +0.163 & +0.001 & +0.135 \\
\bottomrule
\end{tabular}
\end{table}

\begin{table}[t]
\renewcommand{\arraystretch}{1.15}
\centering
\caption{Mean validation error broken down by task-verb family (averaged across the four action dimensions). We omit verb families with fewer than $5$ validation samples.}
\label{tab:verb_breakdown}
\begin{tabular}{lcccc}
\toprule
\textbf{Verb} & OpenVLA & GR00T & $\pi_0$ & SmolVLA \\
\midrule
accompany & 0.300$_{(56)}$ & 2.594$_{(53)}$ & 0.221$_{(56)}$ & 0.197$_{(56)}$ \\
approach & 0.258$_{(229)}$ & 2.595$_{(289)}$ & 0.180$_{(229)}$ & 0.199$_{(229)}$ \\
fly after & 0.665$_{(19)}$ & 3.317$_{(14)}$ & 0.255$_{(19)}$ & 0.245$_{(19)}$ \\
follow & 0.310$_{(108)}$ & 3.349$_{(133)}$ & 0.202$_{(108)}$ & 0.233$_{(108)}$ \\
go after & 0.554$_{(46)}$ & 2.719$_{(62)}$ & 0.264$_{(46)}$ & 0.297$_{(46)}$ \\
move closer & 0.382$_{(52)}$ & 2.383$_{(45)}$ & 0.187$_{(52)}$ & 0.190$_{(52)}$ \\
move towards & 0.247$_{(56)}$ & 2.356$_{(32)}$ & 0.155$_{(56)}$ & 0.148$_{(56)}$ \\
other & 0.370$_{(563)}$ & 2.608$_{(500)}$ & 0.235$_{(563)}$ & 0.268$_{(563)}$ \\
pursuit & 0.422$_{(86)}$ & 2.664$_{(68)}$ & 0.286$_{(86)}$ & 0.336$_{(86)}$ \\
stay close & 0.355$_{(13)}$ & 2.408$_{(12)}$ & 0.193$_{(13)}$ & 0.171$_{(13)}$ \\
\bottomrule
\end{tabular}
\end{table}

\begin{widefigure}[tbp]
  \centering
  \includegraphics[width=\linewidth,height=0.72\textheight,keepaspectratio]{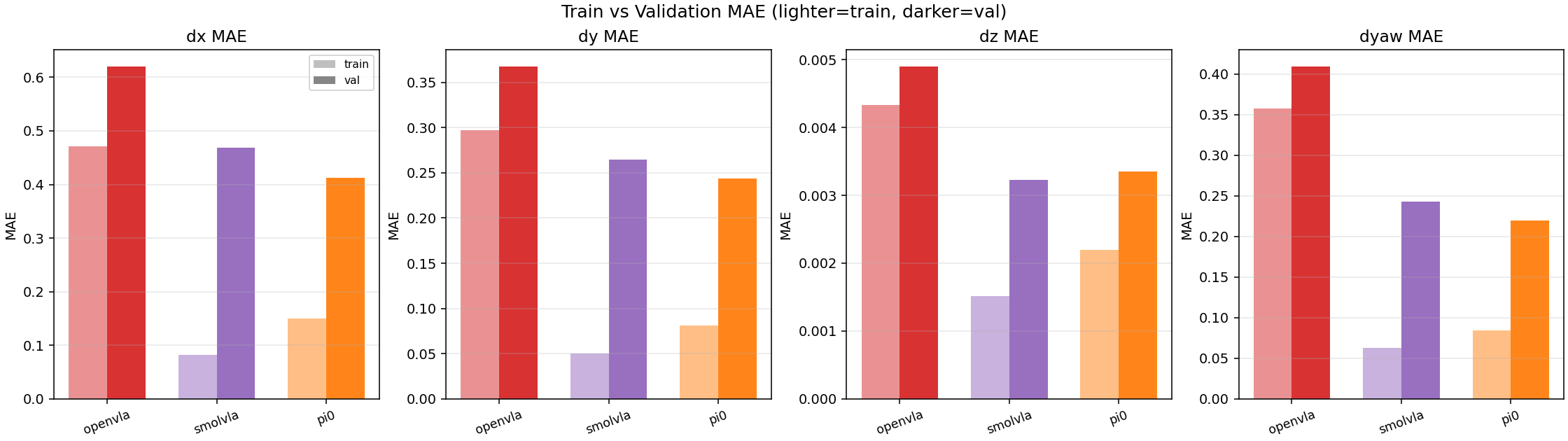}
    \caption{Per-dimension MAE on the train (lighter bars) and validation (darker bars) splits, for each model with train predictions available. Every model shows a large train--val gap; SmolVLA's is the largest despite its lowest training error, while OpenVLA underfits even the training split.}
  \label{fig:train_val}
\end{widefigure}

\subsection{Error vs. action magnitude}

A useful diagnostic for continuous-control policies is how prediction error scales with the ground-truth action magnitude: a policy whose error grows sub-linearly with $|a_{t,d}|$ is more useful for large pursuits than one whose error grows linearly. Fig. \ref{fig:err_mag} bins ground-truth magnitudes into six quantile bins and reports mean validation error (line) and interquartile range (shaded) within each bin, for every model and dimension. On the held-out split, all working models show error that \emph{grows} with $|GT|$ on $\dxl$, the signature of range compression: the policies regress toward the training mean and therefore under-reach the largest expert displacements, so the absolute error is smallest in the low-magnitude bins and largest in the high-magnitude pursuit bins. $\pi_0$ and SmolVLA share this profile with $\pi_0$ slightly flatter (less compression); OpenVLA sits above both with an additional magnitude-independent offset set by its $256$-bin quantization grid. GR00T's error is dimension-uniform and uncorrelated with magnitude. This analysis localizes the dominant error source on the one learnable dimension under-prediction of large displacements and shows it afflicts even the strongest model, underscoring how far the benchmark is from saturation.

\begin{widefigure}[tbp]
  \centering
  \includegraphics[width=\linewidth,height=0.72\textheight,keepaspectratio]{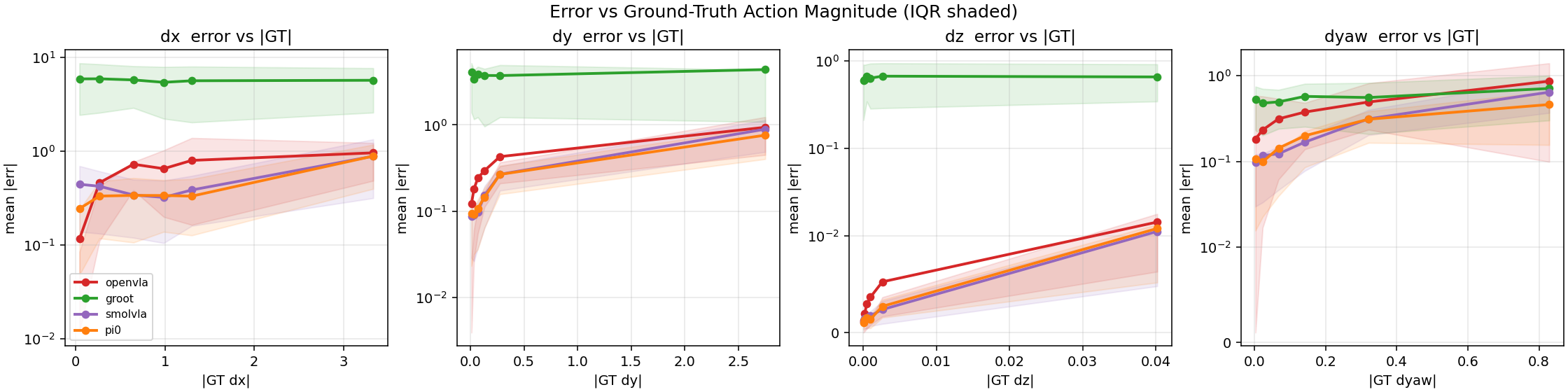}
    \caption{Mean error (line) vs.\ ground-truth magnitude, with interquartile range shaded, binned into six quantile bins. SmolVLA's error grows sub-linearly with $|GT|$ on the in-plane axes; OpenVLA's error is roughly magnitude-independent (constant error floor); GR00T's predictions are essentially decorrelated from $|GT|$.}
  \label{fig:err_mag}
\end{widefigure}

\subsection{Per-episode error heatmap}

Fig. \ref{fig:per_episode_heatmap} aggregates per-episode MAE on a log-color scale across all four models and all four action dimensions. The visualization makes two patterns directly legible. First, the column structure is consistent: the $\pi_0$ and SmolVLA columns are darker (lower error) than OpenVLA, which is in turn far darker than GR00T and this ordering holds across nearly every episode the ranking is not driven by a handful of episodes. Second, there \emph{are} row-aligned ``hard episodes'' that penalize all working models simultaneously (bright rows spanning the $\pi_0$, SmolVLA and OpenVLA columns), predominantly the long approach/locate-subject episodes with large heading changes direct evidence that difficulty is a property of the episode, not merely of the model.

\begin{widefigure}[tbp]
  \centering
  \includegraphics[width=\linewidth,height=0.72\textheight,keepaspectratio]{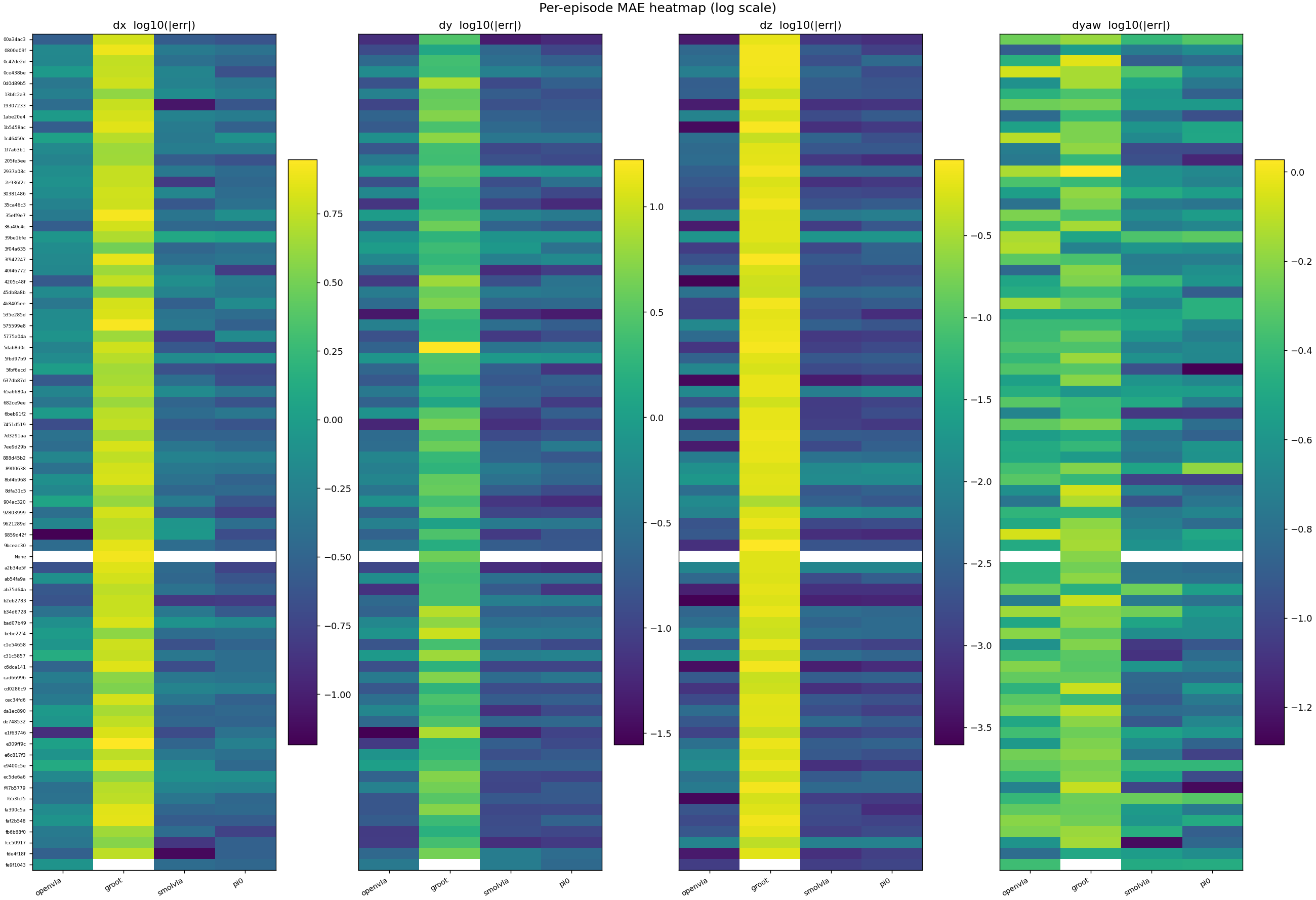}
    \caption{Per-episode MAE heatmap, log-color scaled, with rows corresponding to episodes (shortened UUID labels on the leftmost panel) and columns to models. Each panel shows one action dimension. Darker is lower error; the structural pattern is consistent across dimensions: flow-matching models (\textit{smolvla}, \textit{pi0}) are uniformly low-error while the discrete and diffusion baselines are uniformly high.}
  \label{fig:per_episode_heatmap}
\end{widefigure}

\subsection{Linguistic robustness: Per-verb-family breakdown}

A core motivation for our paraphrase-diverse instruction corpus is to evaluate whether language-conditioned policies are sensitive to the task-initiating verb. Table \ref{tab:verb_breakdown} reports per-verb-family mean validation error for each model, restricted to verb families with at least five held-out samples; under the leakage-free protocol every model is scored on the identical held-out frames, so the columns are directly comparable. Within the working models, error is broadly stable across verb families: $\pi_0$ ranges from $0.155$ (\emph{move towards}) to $0.286$ (\emph{pursuit}), SmolVLA from $0.148$ (\emph{move towards}) to $0.336$ (\emph{pursuit}). The two dominant regimes, \emph{approach} ($N{=}229$) and \emph{follow} ($N{=}108$), yield similar errors ($\pi_0$: $0.180$ and $0.202$; SmolVLA: $0.199$ and $0.233$), indicating the models do not exploit the verb token as a shortcut. Fig. \ref{fig:err_verb} visualizes the same data on a symmetric-log scale. The model ranking ($\pi_0 \approx$ SmolVLA $<$ OpenVLA $\ll$ GR00T) is preserved across all linguistic conditions, so paraphrase diversity is not a hidden confound in the headline comparison.

\begin{widefigure}[tbp]
  \centering
  \includegraphics[width=\linewidth,height=0.72\textheight,keepaspectratio]{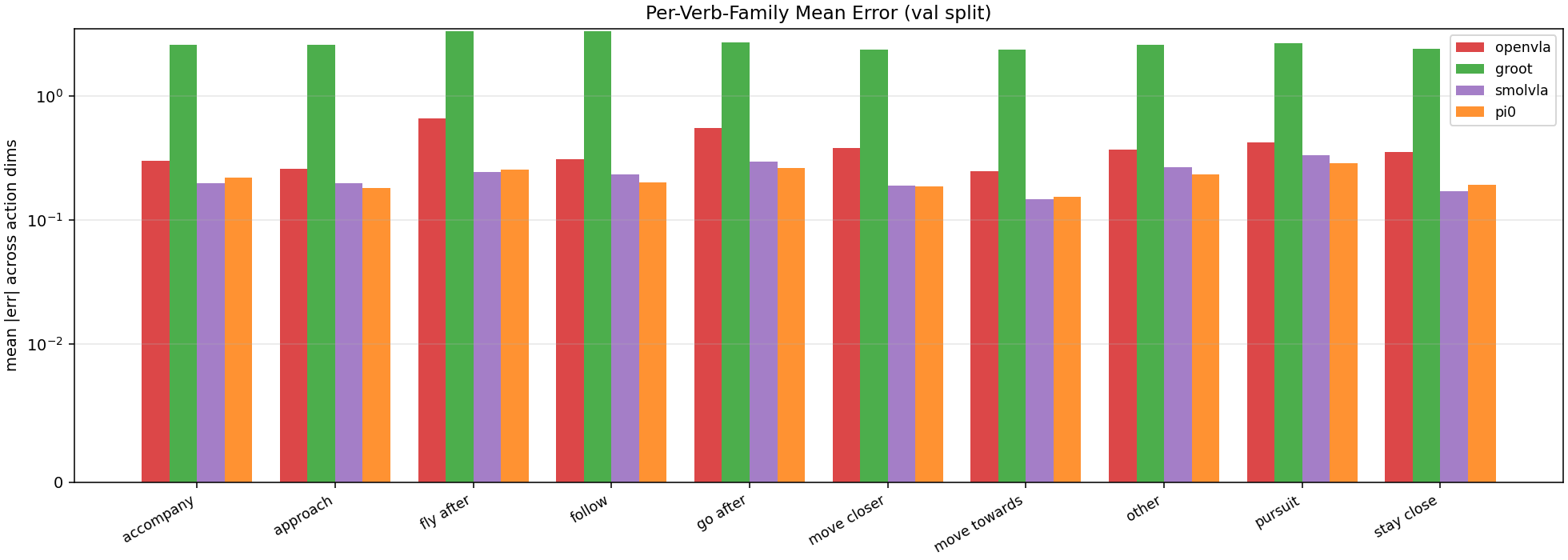}
    \caption{Per-verb-family mean validation error (averaged across the four action dimensions) on a symmetric-log scale. Within each model, performance is stable across verb families; the model ranking is preserved across all linguistic conditions.}
  \label{fig:err_verb}
\end{widefigure}

\subsection{Within-episode error drift}

\begin{figure}[tbp]
  \centering
  \includegraphics[width=\linewidth,height=0.72\textheight,keepaspectratio]{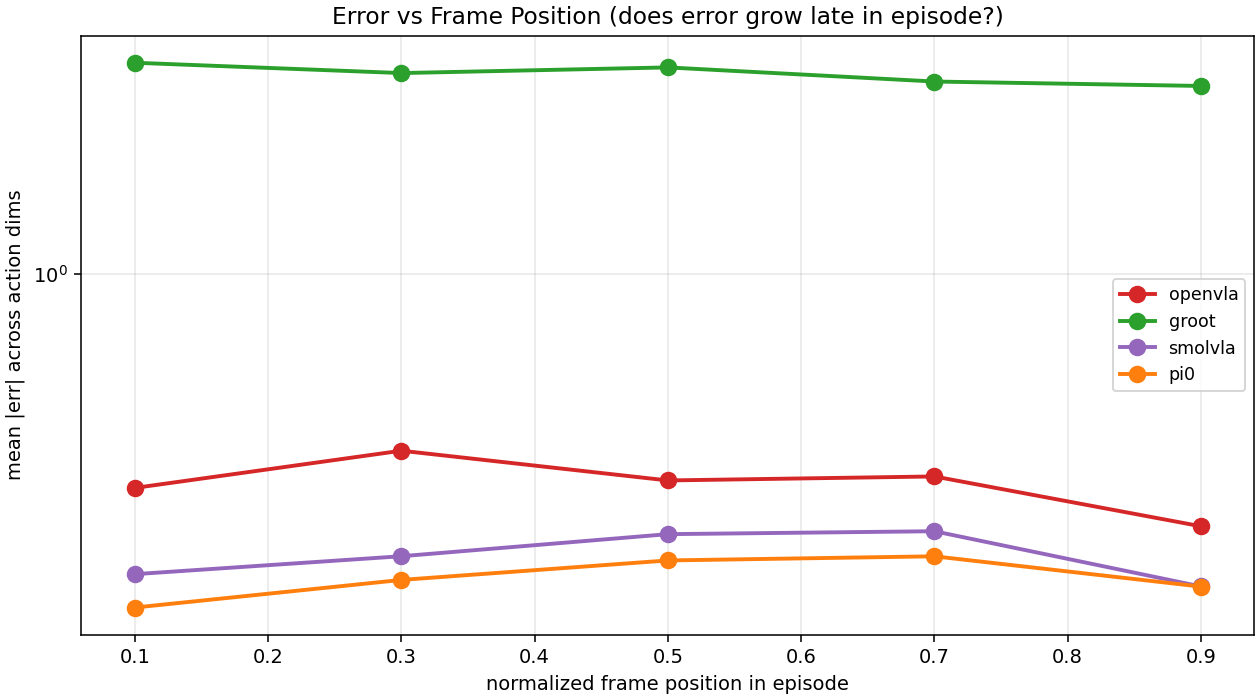}
    \caption{Mean error vs.\ normalized frame position in episode (deciles). All models are essentially flat, indicating no systematic within-episode error drift; this is the expected behavior for a well-curated dataset.}
  \label{fig:err_frame_pos}
\end{figure}

A pure per-step open-loop evaluation cannot directly measure compounding drift, but it can probe a weaker proxy: does the policy degrade \emph{within} an episode as frame index increases? We bin all validation predictions by their normalized position in the parent episode (first decile, second decile, ..., last decile) and report mean error per bin. Fig. \ref{fig:err_frame_pos} shows the result: every model's curve is essentially flat, indicating that errors are uncorrelated with within-episode time. This is consistent with the fact that the trajectories we capture do not contain systematic late-episode events that the policies fail to handle a sanity check for our episode-curation protocol.

\subsection{Side-by-side qualitative comparison}

Fig. \ref{fig:cross_samples} overlays all four models' predictions on the same six validation frames. For each frame we draw the GT in-plane displacement as a solid blue arrow and each model's prediction as a dashed arrow in the model's reference color. The $\pi_0$ and SmolVLA arrows align with the GT arrow in \emph{direction} on most frames but fall short in magnitude, OpenVLA's arrow is shorter still and biased toward the discretized grid centre and GR00T's arrow points far off-frame on nearly every sample. The figure makes the dominant held-out error mode directionally correct but magnitude-compressed in-plane predictions visible at the level of single predictions, while GR00T's gross scale failure is immediately apparent.

\begin{widefigure}[tbp]
  \centering
  \includegraphics[width=\linewidth,height=0.72\textheight,keepaspectratio]{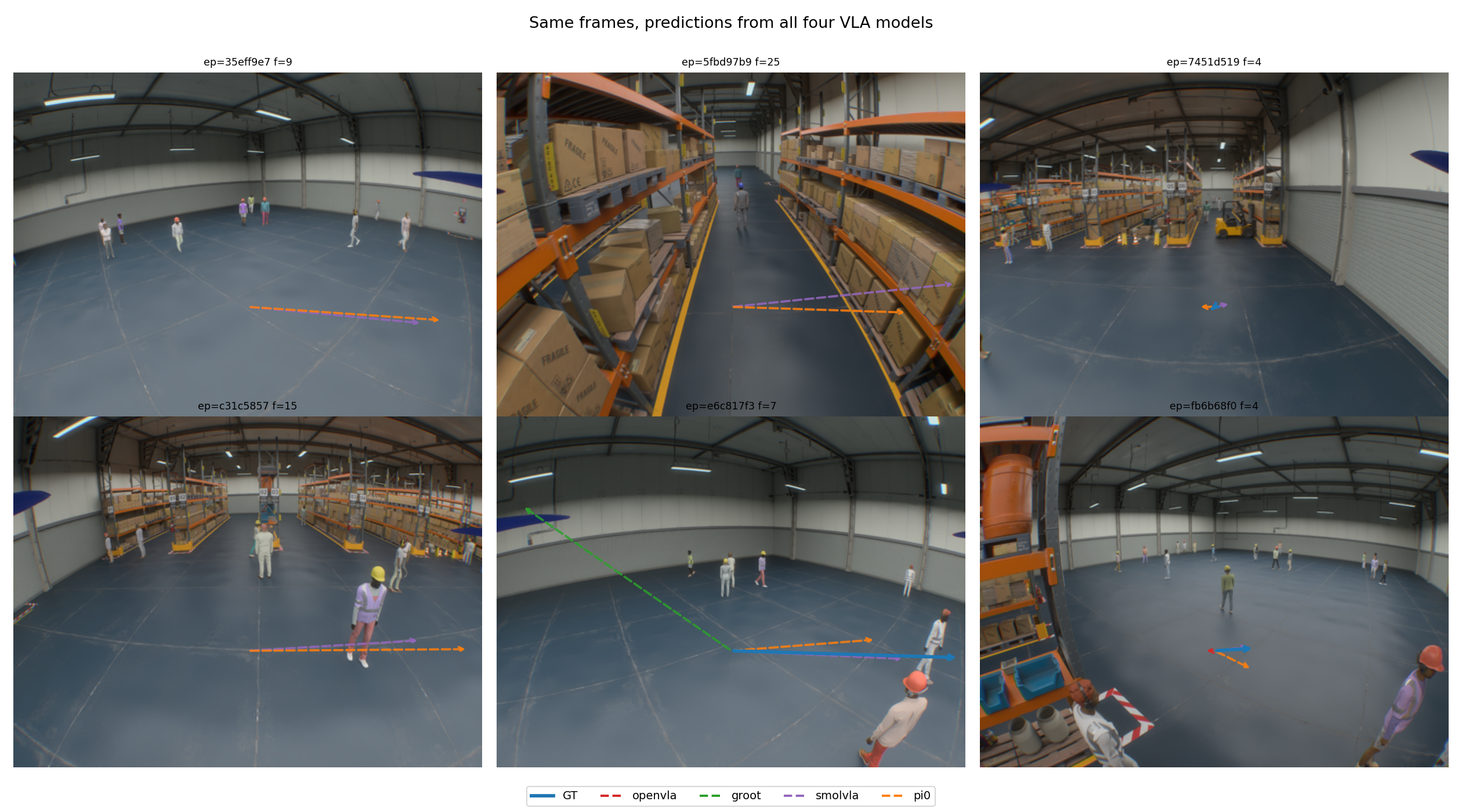}
    \caption{Same six held-out frames, with predictions overlaid from all four VLA models. Solid blue arrows are GT; dashed arrows are per-model predictions in the model's reference color (\textit{red}: OpenVLA; \textit{green}: GR00T; \textit{orange}: $\pi_0$; \textit{purple}: SmolVLA). The flow-matching predictions match GT direction but under-reach in magnitude; the discrete-token prediction undershoots further; the diffusion prediction points off-frame.}
  \label{fig:cross_samples}
\end{widefigure}

\section{Extended Experimental Results}
\label{app:extended}
This appendix collects supplementary diagnostics that corroborate the headline analysis but are too fine-grained for the main text. Figures are grouped by theme; unless noted, $1$ FPS plots span all four baselines and $10$ FPS plots cover the accuracy frontier ($\pi_0$, SmolVLA).

\subsection{Correlation and per-episode ranking summaries}
\begin{figure}[tbp]
\centering
\includegraphics[width=\linewidth,height=0.72\textheight,keepaspectratio]{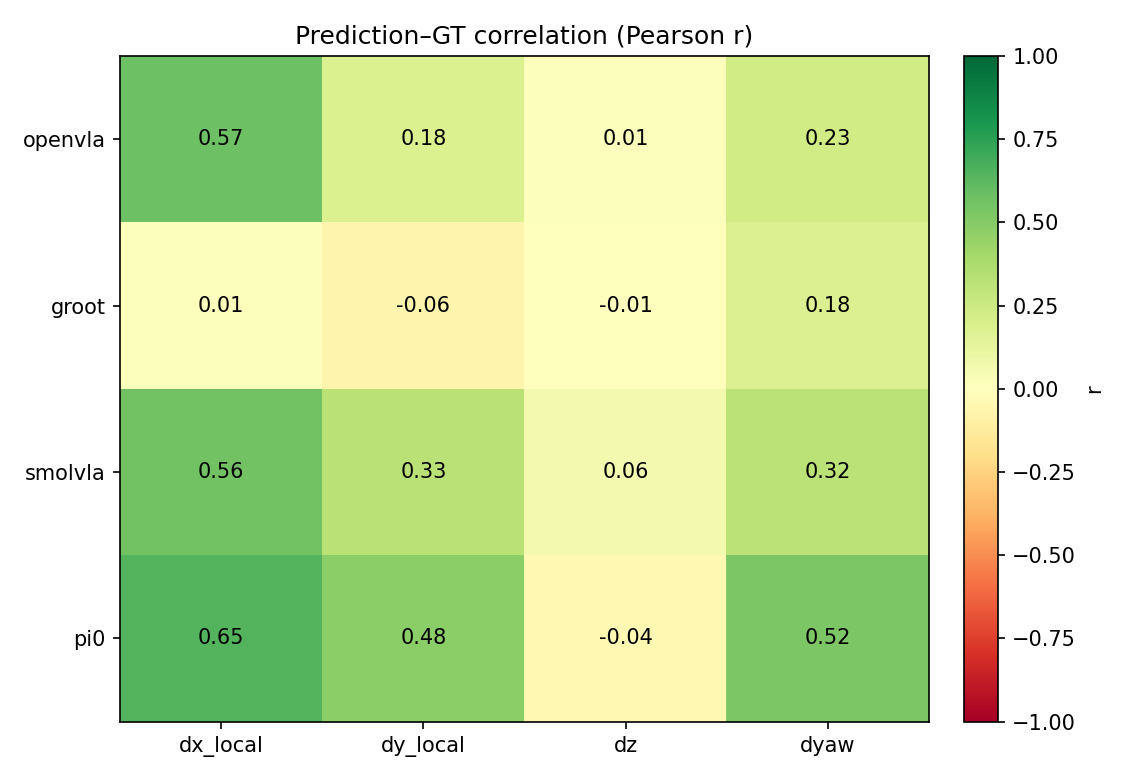}
\caption{Per-dimension prediction--ground-truth Pearson correlation for all four baselines at $1$ FPS. Greener is higher. The forward channel is the most learnable for every model; $\dz$ is at chance throughout; $\pi_0$ dominates column-wise.}
\label{fig:r_heatmap_app}
\end{figure}

\begin{figure}[tbp]
\centering
\includegraphics[width=\linewidth,height=0.72\textheight,keepaspectratio]{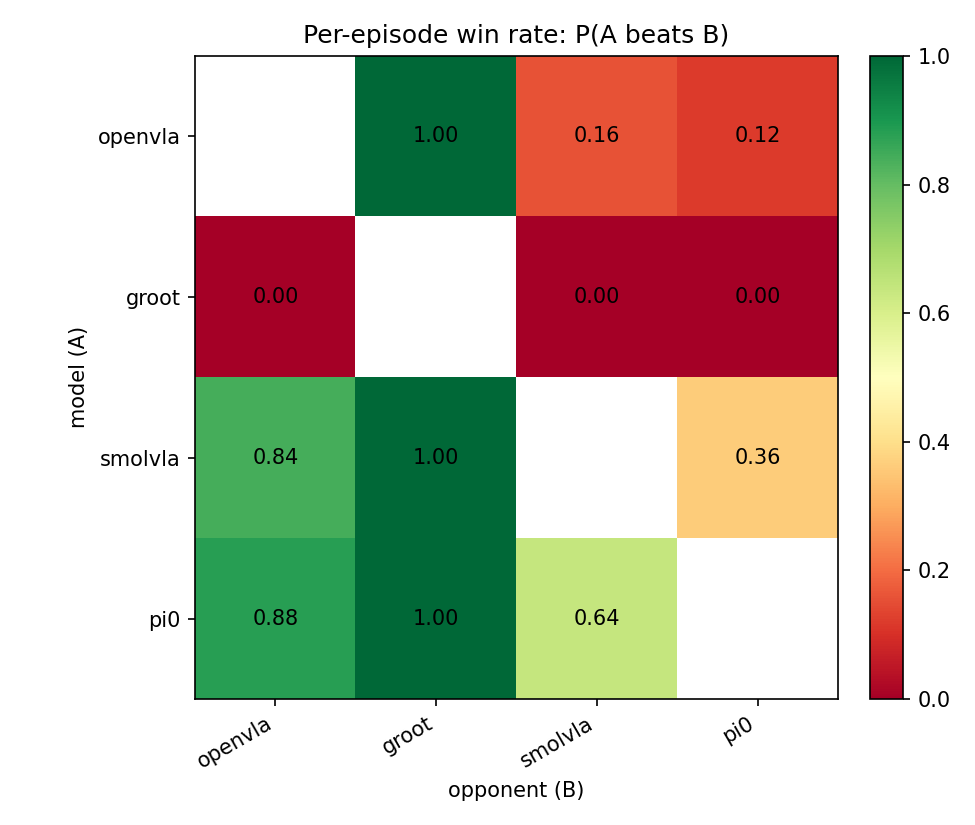}
\caption{Pairwise per-episode win rate $P(\text{row beats column})$ in mean held-out MAE at $1$ FPS. The ordering $\pi_0 \!>\!$ SmolVLA $\!>\!$ OpenVLA $\!>\!$ GR00T holds at the per-episode level, not merely in aggregate.}
\label{fig:winrate_app}
\end{figure}

Fig. \ref{fig:r_heatmap_app} compresses the per-dimension Pearson correlation of all four models at $1$ FPS into a single heatmap, making the learnability hierarchy ($\dxl{>}\dpsi{>}\dyl{>}\dz$) and the model ordering legible at a glance. Fig. \ref{fig:winrate_app} reports the pairwise per-episode win rate $P(A \text{ beats } B)$, confirming that the ranking is a per-episode property rather than an averaging artifact: $\pi_0$ beats SmolVLA on $64\%$ of episodes, both flow-matching models beat OpenVLA on $\geq 84\%$ and every model beats GR00T on $100\%$.

\subsection{Directional accuracy and calibration at 10 FPS}
The underdispersion and directional-accuracy signatures discussed in the main text are visualized for each model in Figs. \ref{fig:calib_app}--\ref{fig:yaw_app}. The speed-calibration histograms (Fig. \ref{fig:calib_app}) show the predicted horizontal-step magnitude distribution shifted below ground truth (mean pred/GT $\approx$ $0.85$ for $\pi_0$), the magnitude-compression signature. The yaw-turn confusion matrices (Fig. \ref{fig:yaw_app}) show that ``straight'' is recovered reliably while left/right turns are frequently predicted as straight the directional counterpart of the modest heading correlation.

\begin{widefigure}[tbp]
\centering
\includegraphics[width=0.49\linewidth,height=0.72\textheight,keepaspectratio]{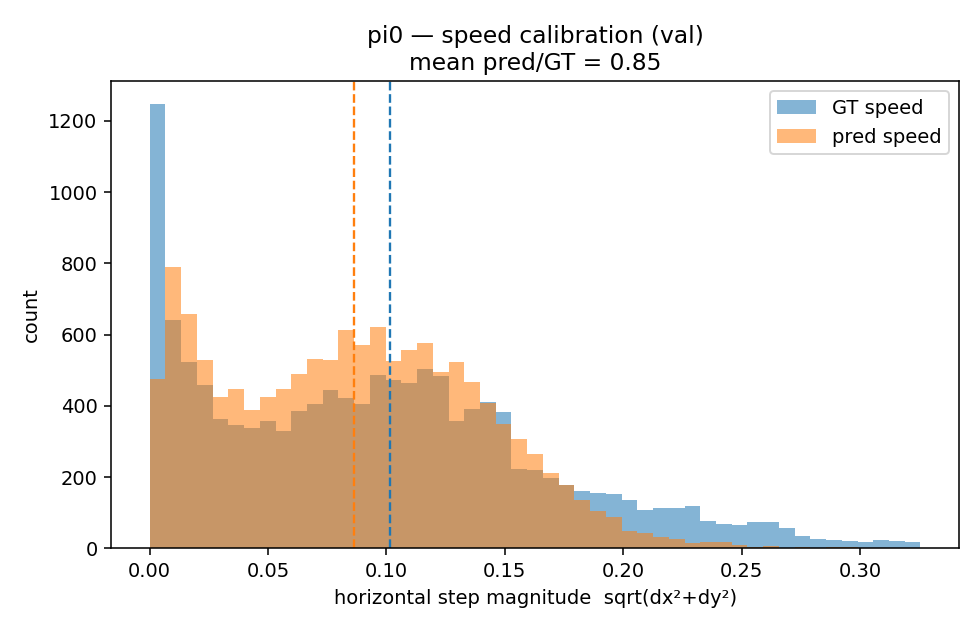}\hfill
\includegraphics[width=0.49\linewidth,height=0.72\textheight,keepaspectratio]{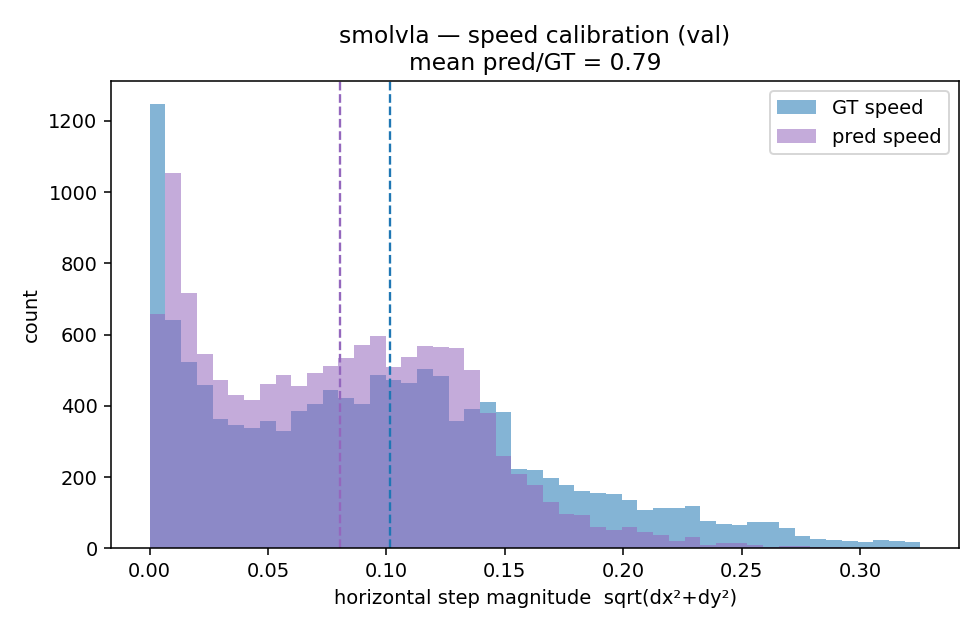}
\caption{Horizontal-speed calibration at $10$ FPS: ground-truth (blue) vs.\ predicted (orange) step-magnitude distributions for $\pi_0$ (left) and SmolVLA (right). Both models under-predict the spread (dashed means), the magnitude-compression signature.}
\label{fig:calib_app}
\end{widefigure}

\begin{widefigure}[tbp]
\centering
\includegraphics[width=0.49\linewidth,height=0.72\textheight,keepaspectratio]{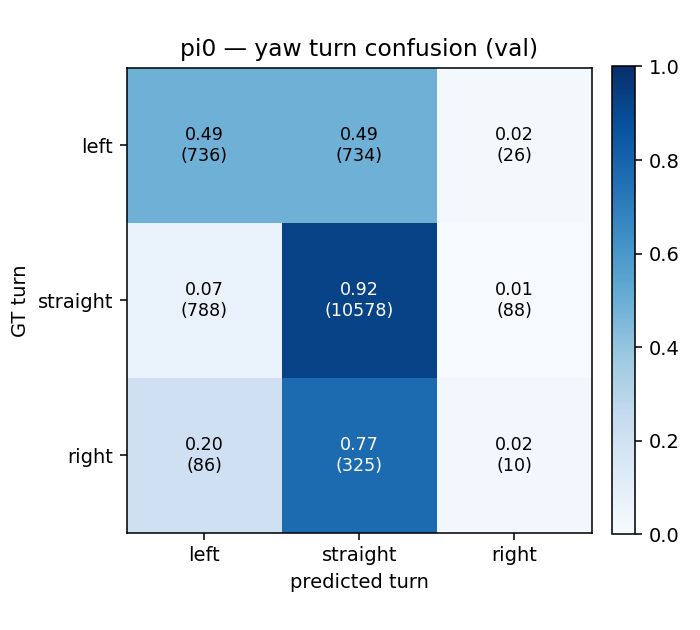}\hfill
\includegraphics[width=0.49\linewidth,height=0.72\textheight,keepaspectratio]{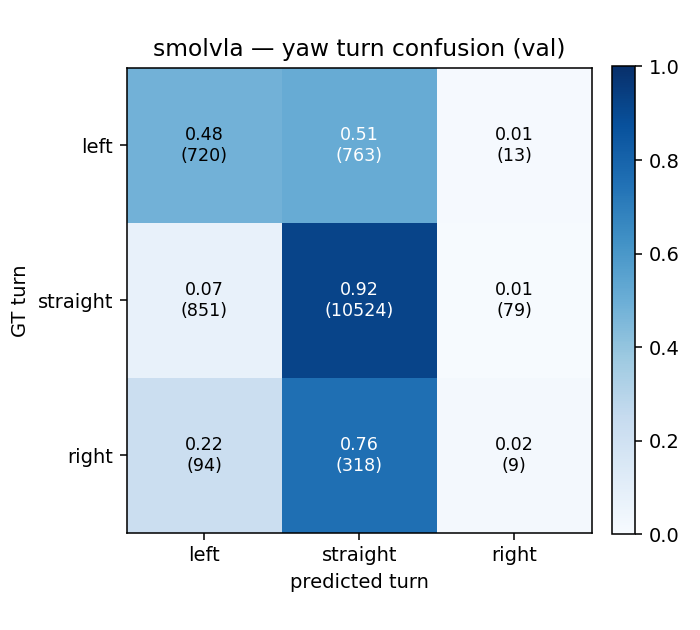}
\caption{Yaw-turn confusion matrices at $10$ FPS (rows: ground-truth turn; columns: predicted) for $\pi_0$ (left) and SmolVLA (right). Straight motion is recovered well; left/right turns collapse toward ``straight'', explaining the modest $\dpsi$ correlation.}
\label{fig:yaw_app}
\end{widefigure}

\subsection{SmolVLA companion diagnostics and the 10 FPS error CDF}
\begin{figure}[tbp]
\centering
\includegraphics[width=\linewidth,height=0.72\textheight,keepaspectratio]{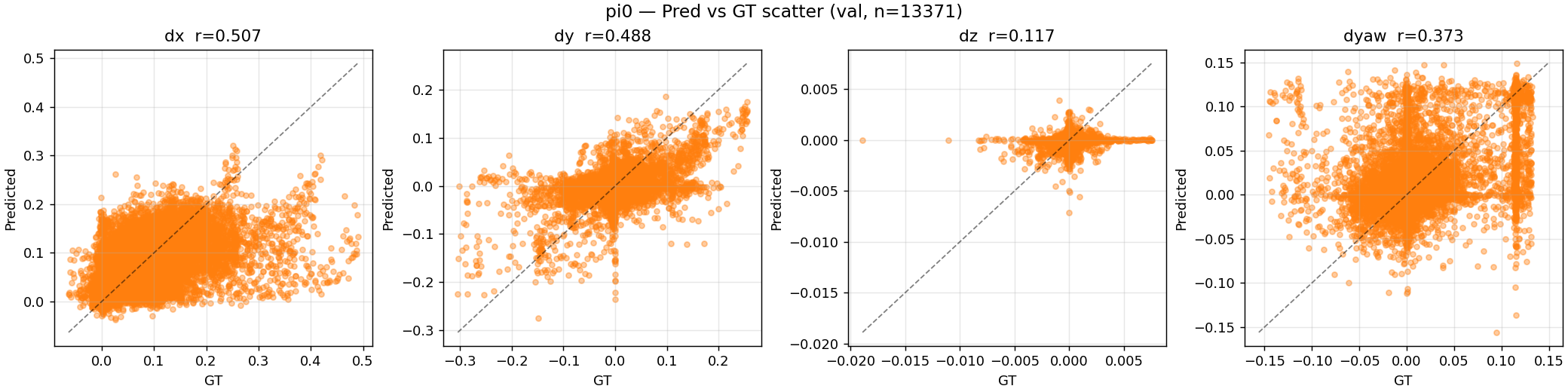}
\caption{Per-dimension prediction-vs.-ground-truth scatter for $\pi_0$ on the $10$ FPS held-out split ($N{=}13{,}371$). The dashed diagonal is the identity line. The qualitative signature is identical to $1$ FPS (Fig. \ref{fig:scatter_pi0}) at one-tenth the magnitude scale: a positive but range-compressed $\dxl$ trend, partial $\dyl$ structure and $\dz$ collapsed to the prior band.}
\label{fig:scatter_pi0_10fps}
\end{figure}

For completeness, Fig. \ref{fig:smolvla_scatter_app} gives SmolVLA's per-sample scatter at $10$ FPS (the companion to $\pi_0$'s Fig. \ref{fig:scatter_pi0_10fps}) and Fig. \ref{fig:smolvla_traj_app} its open-loop rollouts, both showing the same range-compression and curve-drift behavior as $\pi_0$ but slightly more pronounced. Fig. \ref{fig:cdf_10fps_app} plots the per-dimension absolute-error CDF at $10$ FPS, the cumulative view of the per-step MAE saturation discussed in Section \ref{sec:results}.

\begin{widefigure}[tbp]
\centering
\includegraphics[width=\linewidth,height=0.72\textheight,keepaspectratio]{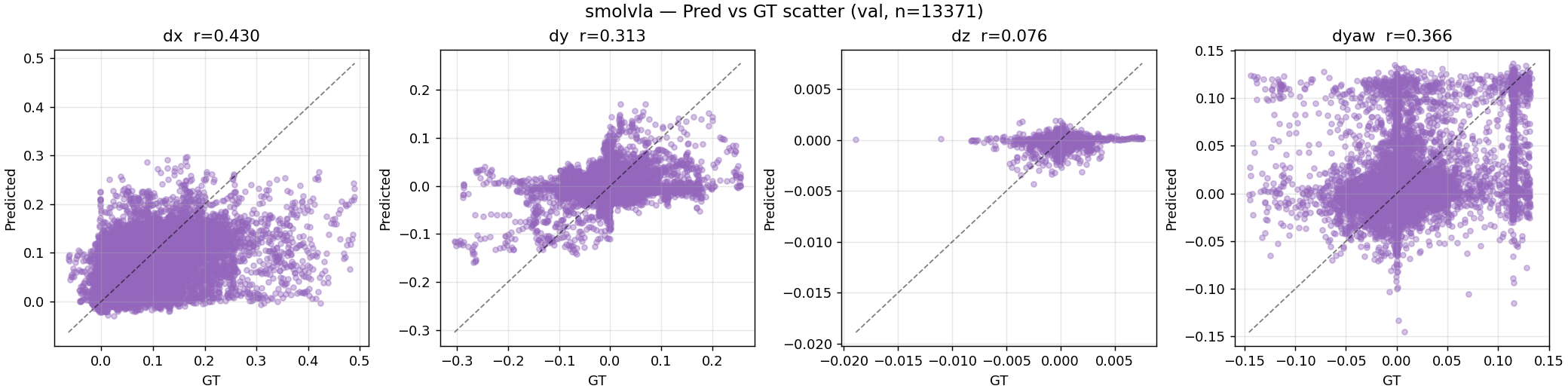}
\caption{SmolVLA per-dimension prediction-vs.-ground-truth scatter at $10$ FPS. The pattern matches $\pi_0$ (Fig. \ref{fig:scatter_pi0_10fps}) with steeper $\dxl$ range compression, consistent with SmolVLA's larger generalization gap.}
\label{fig:smolvla_scatter_app}
\end{widefigure}

\begin{widefigure}[tbp]
\centering
\includegraphics[width=\linewidth,height=0.72\textheight,keepaspectratio]{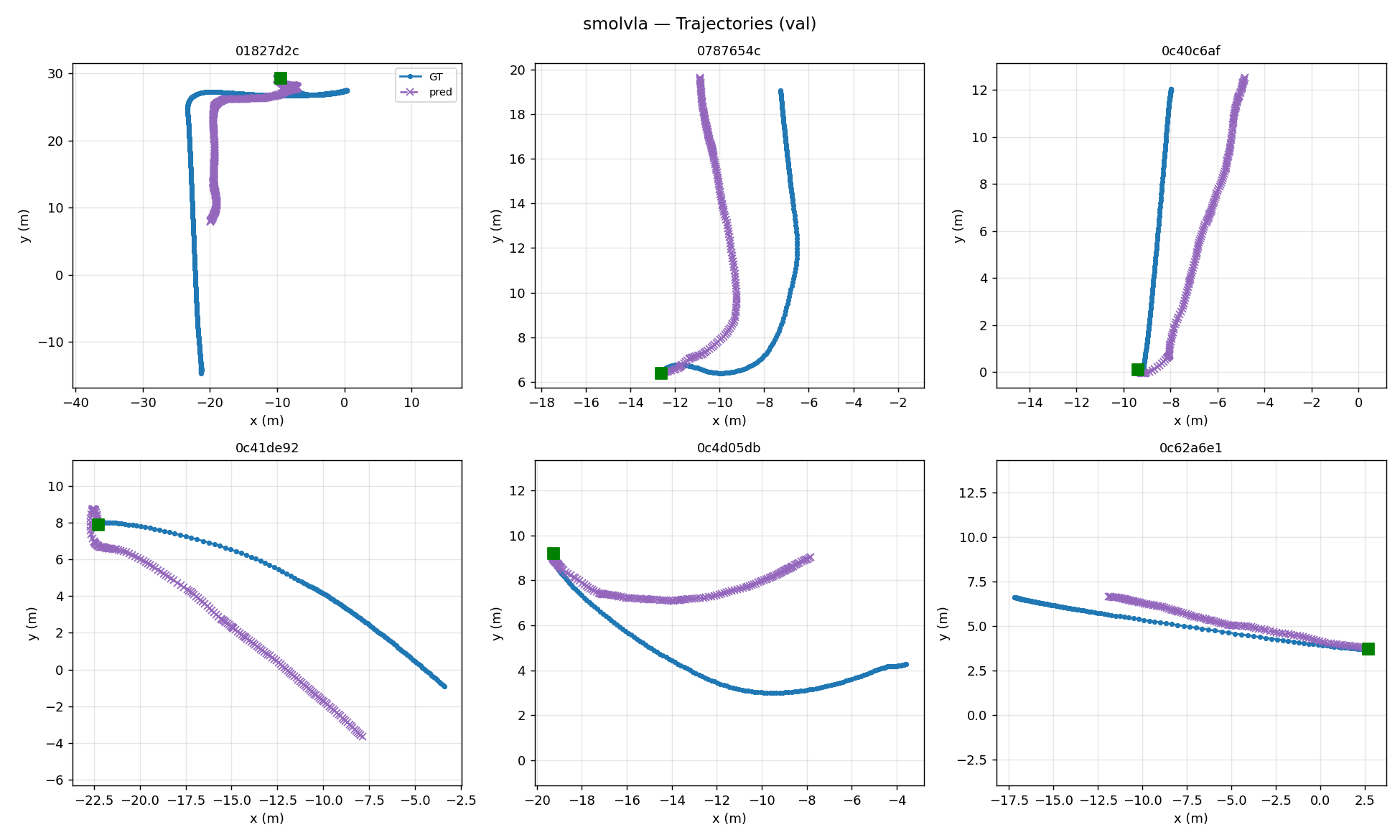}
\caption{SmolVLA open-loop rollouts (orange) vs.\ expert paths (blue) on six $10$ FPS held-out episodes. Drift on curved segments is larger than $\pi_0$'s (Fig. \ref{fig:traj_10fps}), consistent with the ADE/FDE ranking of Table \ref{tab:rollout}.}
\label{fig:smolvla_traj_app}
\end{widefigure}

\begin{widefigure}[tbp]
\centering
\includegraphics[width=\linewidth,height=0.72\textheight,keepaspectratio]{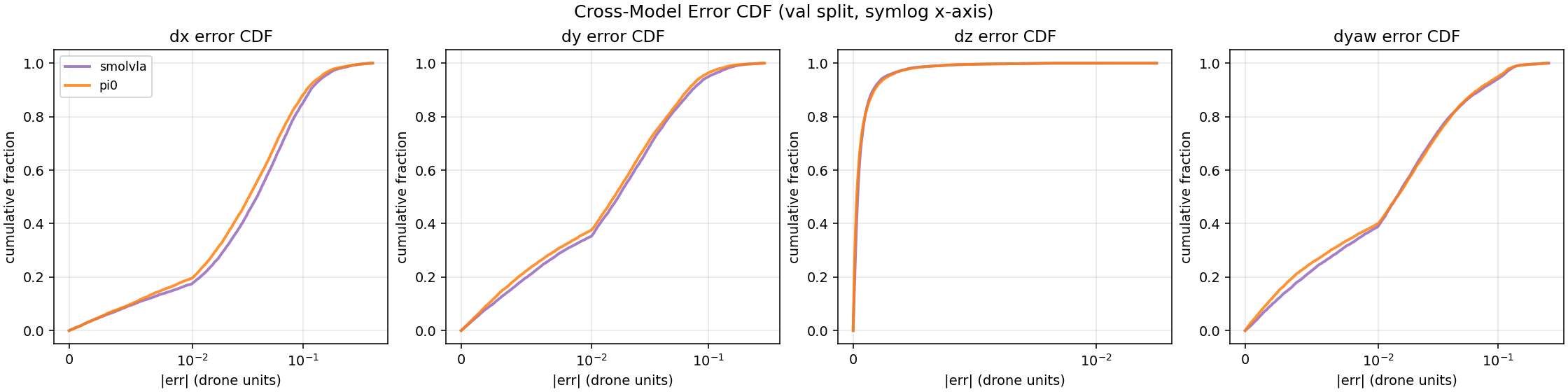}
\caption{Per-dimension absolute-error CDF at $10$ FPS (symmetric-log $x$). $\pi_0$ holds the right-tail edge over SmolVLA on $\dxl$, while both nearly coincide with the naive baseline on $\dyl$/$\dz$/$\dpsi$ the cumulative view of the per-step MAE saturation at high control rate.}
\label{fig:cdf_10fps_app}
\end{widefigure}

\bibliographystyle{elsarticle-num}
\bibliography{warefly_refs}

\end{document}